\PassOptionsToPackage{table}{xcolor}
\documentclass{article}

\usepackage{xcolor}
\usepackage[preprint]{corl_2026}

\usepackage{microtype}
\usepackage{booktabs}
\usepackage{multirow}
\usepackage{array}
\usepackage{graphicx}
\usepackage{algorithm}
\usepackage{algorithmicx}
\usepackage{algpseudocode}
\usepackage{listings}
\usepackage{tabularx}
\usepackage{subcaption}
\usepackage{wrapfig}
\usepackage{placeins}
\usepackage{pgffor}
\usepackage{makecell}
\usepackage{amsmath,amssymb,amsthm,bm,mathtools}
\usepackage[most]{tcolorbox}
\usepackage{enumitem}
\usepackage{siunitx}
\usepackage[capitalize,noabbrev]{cleveref}
\usepackage{xspace}

\hypersetup{
  pdftitle={Robust Bimanual Vision-Language-Action Models via Embarrassingly Simple Modality Masking},
  pdfauthor={Dongzhou Cheng, Ziang Li, Yixiao Zhou, Haojuan Li, Jinghao Zhang, Lei Lei, Minjing Dong, Jie Gui, Jiaqi Wang}
}

\providecommand{\Description}[1]{}

\newcommand{\eg}{e.g.\@\xspace}

\newtcolorbox{principlebox}[1][]{
  enhanced, breakable,
  colback=white, colframe=black!10,
  boxrule=0.6pt, arc=2pt,
  left=8pt, right=8pt, top=8pt, bottom=8pt,
  attach boxed title to top left={yshift=-2pt, xshift=6pt},
  boxed title style={frame code={
      \path[draw=black!10, fill=black!3, rounded corners=1pt]
      ([yshift=-1pt]frame.north west) rectangle
      ([xshift=0pt,yshift=6pt]frame.north east);
    }, interior engine=empty},
  title=\textbf{Design Principles for M3},
  #1
}

\title{Robust Bimanual Vision-Language-Action Models via Embarrassingly Simple Modality Masking}

\author{
  \normalsize
  \textbf{Dongzhou Cheng$^{1,2}$, Ziang Li$^{3,2}$, Yixiao Zhou$^{4,2}$, Haojuan Li$^{6,2}$} \\
  \textbf{Jinghao Zhang$^{5,2}$, Lei Lei$^{5,2}$, Minjing Dong$^{7}$, Jie Gui$^{1}$, Jiaqi Wang$^{2}$} \\[2mm]
  \small
  $^1$Southeast University \quad $^2$Shanghai Innovation Institute \quad $^3$Wuhan University \\
  $^4$Zhejiang University \quad $^5$University of Science and Technology of China \\
  $^6$Shanghai Jiao Tong University \quad $^7$City University of Hong Kong \\[2mm]
  \href{https://m3vla.github.io/}{\texttt{m3vla.github.io}}
}

\begin{document}

\maketitle

\begin{abstract}
Query-based Vision-Language-Action (VLA) models offer low-latency inference that is attractive for bimanual robotic manipulation, but we observe that they can still exhibit discontinuous actions and execution failures in complex dual-arm tasks. We hypothesize that unstable multi-view and language fusion is one contributing factor in these failures, often coinciding with attention spreading to distracting regions. To improve robustness, we introduce the Modality Masking Mechanism (M3), an \emph{embarrassingly simple}, training-only strategy that requires no architectural changes or large-scale robot pretraining. M3 stochastically masks subsets of modality channels during training, exposing the policy to controlled partial observations and encouraging it to rely less on distracting cues and more on evidence that remains reliable. We evaluate M3 on ten bimanual tasks from RoboTwin 2.0 and on three long-horizon real-world tasks. Compared with the Adapter baseline, M3 improves average success by 21.7\% in the Clean setting and 11.4\% in Clean2Rand, where policies are trained on clean demonstrations and evaluated on randomized scenes, while also improving averaged real-world full-task success by over 30\%. These results suggest that structured training-time masking is a practical way to improve the robustness of query-based VLA policies for bimanual manipulation.
\end{abstract}


\section{Introduction}
\label{sec:intro}
Bimanual robotic manipulation is widely viewed as a key stepping stone toward general-purpose embodied agents, since it demands coordinated dual-arm control~\cite{zhao2023learning,fu2024mobile,chen2025robotwin}, contact-rich interaction, and precise spatiotemporal reasoning. As Vision-Language-Action (VLA) models~\cite{zitkovich2023rt,kim2024openvla,kim2025fine,black2024pi_0,liu2024rdt,zhong2025survey,shao2025large,bjorck2025gr00t,intelligence2025pi_} progress from passive multimodal understanding of vision-language models (VLMs)~\cite{beyer2024paligemma,karamcheti2024prismatic,bai2025qwen2} to acting in the physical world, efficient real-time inference becomes essential for bimanual control. Among existing paradigms, query-based VLA~\cite{kim2025fine,wang2025vla} architectures decode actions with parallel learnable queries in a single forward pass, offering natural parallelism and low-latency inference.

\begin{figure}[t!]
    \centering
    \includegraphics[width=0.70\linewidth]{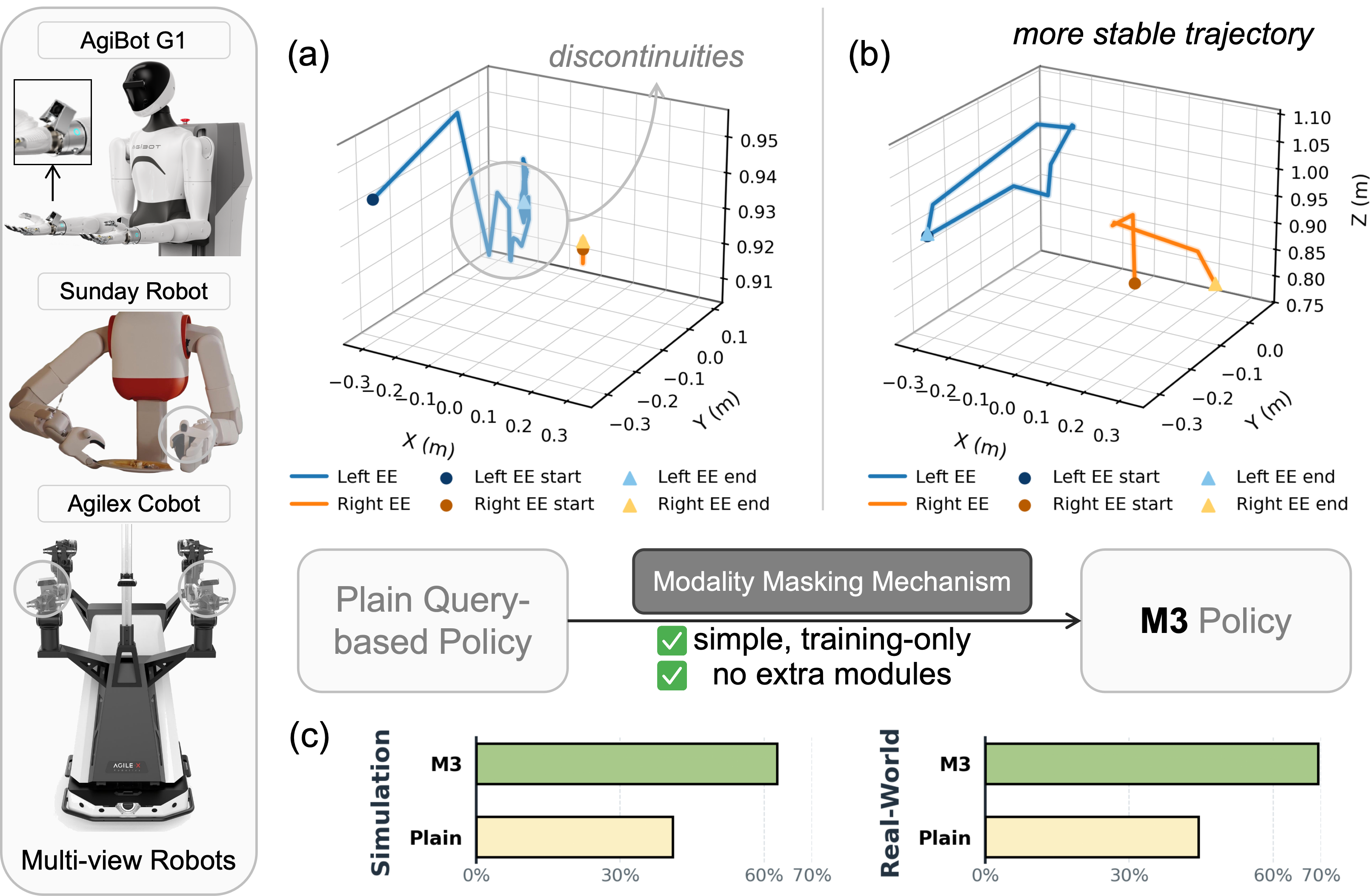}
    \caption{Left: representative multi-view, dual-arm platforms that reflect a prevalent form factor in current embodied manipulation research. (a) A plain query-based VLA exhibits trajectory discontinuities near the contact region. (b) The M3-trained policy produces more stable end-effector paths and more consistent contact alignment. (c) As shown in the bottom bar charts, M3 yields consistent improvements over the plain baseline across both simulation and real-world evaluations under clean conditions.}
    \Description{Two 3D end-effector trajectory plots compare a plain query-based VLA with M3. The baseline trajectory is jittery and discontinuous near contact, while the M3 trajectory is smoother and remains aligned with the intended contact path.}
    \label{fig:smooth}
\end{figure}

\begin{figure}[t!]
  \centering
  \includegraphics[width=.70\linewidth]{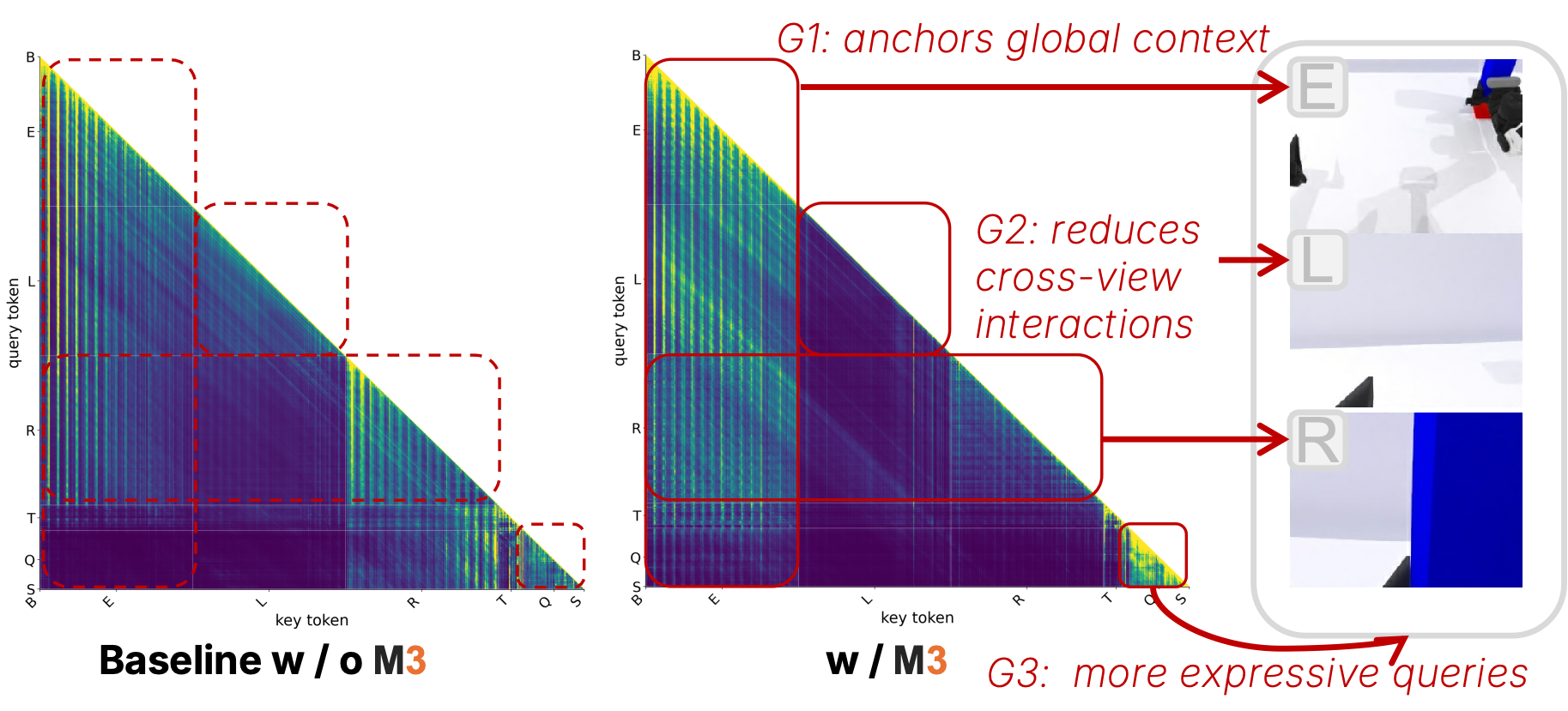}
  \par\vspace{.45em}
  \includegraphics[width=.70\linewidth]{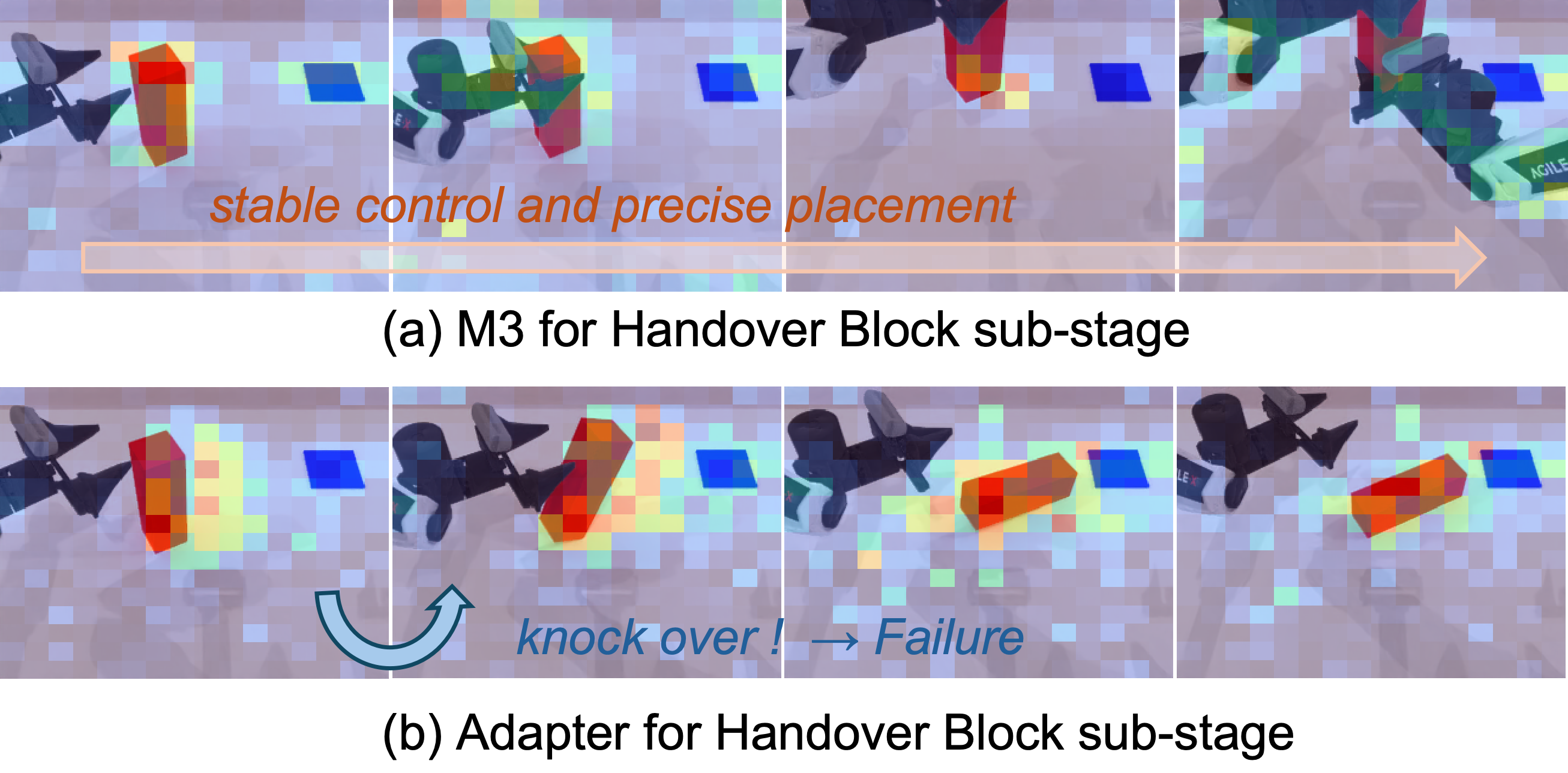}
  \caption{\textbf{Attention structure and contact-rich rollout behavior.}
  \textbf{Top:} Post-hoc attention-score maps for the Adapter baseline without M3 (left) and M3 (right). Rows are query tokens and columns are key tokens; E, L, R, T, and Q denote the egocentric, left-wrist, right-wrist, text, and action-query token groups, respectively. The M3 map exhibits patterns consistent with anchoring global context in the egocentric stream, reducing cross-view shortcuts, and encouraging more expressive query interactions.
  \textbf{Bottom:} At the handover-to-placement stage of the long-horizon \textit{Handover Block} task, M3 maintains stable dual-arm coordination and completes placement in the shown rollout, whereas the Adapter baseline knocks the block over during contact and does not recover. Together, the panels juxtapose the intended change in attention structure with the qualitative execution difference observed in a representative rollout.}
  \Description{A vertically stacked figure pairs attention-score maps with a Handover Block rollout. The top compares the Adapter baseline without M3 against M3 across egocentric, wrist-camera, text, and action-query token groups. The bottom shows M3 completing handover and placement while the baseline knocks over the block and fails to recover.}
  \label{fig:mechanism_handover}
\end{figure}

However, we observe that query-based VLAs can still exhibit discontinuous actions in bimanual tasks, leading to poor execution (see Fig.~\ref{fig:smooth}). In many failure cases, the model's attention spreads over irrelevant image regions rather than remaining concentrated on the intended object or contact region: multi-view inputs introduce salient yet task-irrelevant activations, which we find are often associated with unstable action rollout and reduced success rates (Fig.~\ref{fig:heatmap}, Sec.~\ref{sec:analysis}).

\paragraph{Why does attention misalignment occur, and how can masking help?}

We observe a recurring failure pattern: when all camera views are always visible during training, the policy learns spurious cross-view correlations rather than reasoning about which view provides reliable evidence for the current action. For instance, if a distractor in the idle arm's view has high visual saliency, the model may incorrectly attend to it even when planning actions for the active arm. This manifests as diffuse attention spreading (Fig.~\ref{fig:heatmap}) and severe degradation under distribution shift, where every method we evaluate drops sharply (Table~\ref{tab:clean2rand_performance_comparison}): the Adapter baseline falls from 41.0\% average success in Clean (Table~\ref{tab:performance_comparison}) to 3.7\%, and the strongest pretrained policy reaches only 12.9\%. Our aim is to reduce this gap rather than to close it.

Why does this happen? Standard training exposes the model to all views simultaneously, allowing it to exploit any correlation—including unstable ones such as spurious visual feature matching across views or reliance on idle-arm saliency cues that happen to correlate with success in clean demonstrations. These correlations work when the training distribution is narrow but break when distractors, occlusions, or lighting changes alter one view's appearance. The model never learns which cues remain informative under perturbation and which are coincidental to the clean training scenes.

Drawing on decision-making under partial observability~\cite{kaelbling1998planning,lauri2022partially}, we propose training under \emph{structured view dropout}: masking the wrist-view group while preserving the egocentric view as a stable spatial reference forces the policy to solve tasks using evidence that remains reliable across different masking patterns. When the wrist views are jointly masked, the model cannot rely on wrist-to-ego saliency matching; it must instead extract task geometry from the egocentric frame, and learn to use wrist evidence only when those views provide non-redundant local information. We formalize this via the \textbf{Modality Masking Mechanism (M3)}, guided by three bimanual-specific design principles: (1)~preserve the egocentric view as a consistent spatial anchor, (2)~mask dual-wrist views \emph{jointly} to prevent idle-arm interference, and (3)~apply query-subset masking to encourage complementary action representations. Fig.~\ref{fig:mechanism_handover} illustrates the resulting difference: the top panel contrasts the post-hoc attention structure of the Adapter baseline and M3, while the bottom panel shows their execution in the same \textit{Handover Block} stage. 

M3 is an \emph{embarrassingly simple}, training-only intervention: it requires no architectural changes or large-scale pretraining, making it a low-overhead strategy for improving robustness. We evaluate on the challenging \textbf{RoboTwin 2.0}~\cite{chen2025robotwin} benchmark that spans diverse bimanual tasks and provides both Clean and Clean2Rand evaluation settings, as well as on three long-horizon real-world tasks. Clean2Rand trains policies on clean demonstrations and evaluates them on randomized scenes. With the same backbone and tuning budget as the Adapter baseline~\cite{wang2025vla}, M3 improves average success by \textbf{21.7\%} in Clean and by \textbf{11.4\%} in Clean2Rand. On the real robot, M3 raises the averaged full-task success rate by \textbf{25.0\%} under clean conditions and by \textbf{48.6\%} under OOD clutter across three long-horizon tasks. Overall, our \textbf{Contribution} lies in identifying a structured visibility design tailored to the bimanual multi-view setting, which can serve as a practical training-time strategy for improving the robustness of query-based VLA policies for bimanual robotic manipulation.

\begin{figure*}[t!]
  \centering
  \includegraphics[width=.95\textwidth]{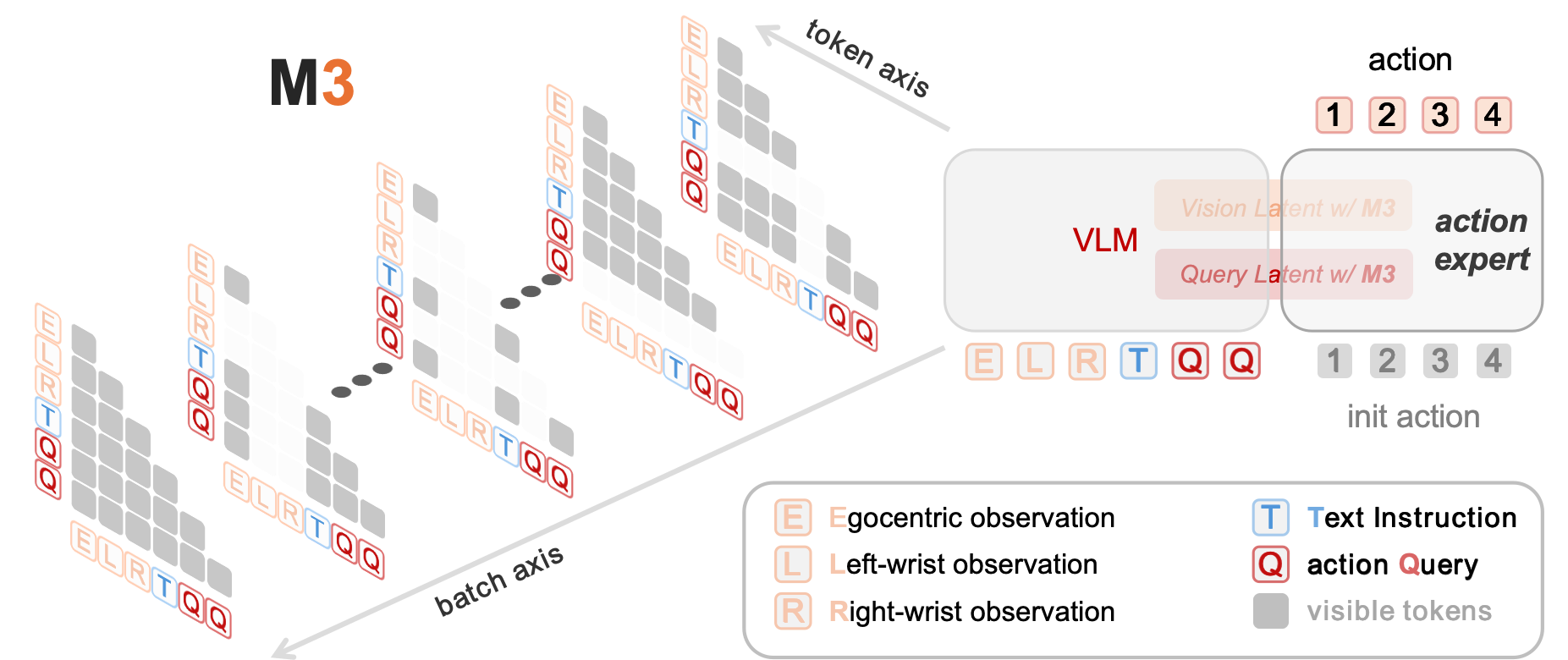}
\caption{Overview of the Modality Masking Mechanism (M3). M3 is a training-time strategy that injects partial observability by stochastically masking modalities and action queries. As illustrated along the batch and token axes, different token subsets (comprising Egocentric (E), Left-wrist (L), Right-wrist (R) observations, Text Instructions (T), and Action Queries (Q)) are made visible (grey) at each training step. These partially observable inputs are processed by the VLM to produce M3-conditioned latents, which are then used by the action expert to decode actions. This process is intended to encourage more robust evidence selection and complementary representations without over-relying on any single input channel.}
  \Description{A schematic overview of M3 shows token groups for the egocentric camera, left and right wrist cameras, language tokens, and action queries. Different subsets are stochastically masked during training, and the partially visible tokens are passed through the VLM and action expert to produce action predictions.}
  \label{fig:overview}
\end{figure*}

\section{Preliminaries}
\label{sec:preliminaries}

We consider \emph{query-based} VLA policies~\cite{kim2025fine,wang2025vla} that turn multi-view images and language into an $H$-step action chunk in a single forward pass. For simplicity, we treat the proprioceptive inputs available during training as the default inputs to the VLA and omit further detailed description. At time $t$, let $V_t = \big[\phi_v(X_t^{(1)}), \ldots, \phi_v(X_t^{(K)})\big] \in \mathbb{R}^{N_v \times d}$ denote visual tokens from $K$ cameras, $T = \phi_\ell(P) \in \mathbb{R}^{N_p \times d}$ the language tokens for instruction $P$, and $Q \in \mathbb{R}^{N_q \times d}$ the decoder's $N_q$ action queries. A generic model $\mathbf{g}_\theta$ predicts
\begin{equation}
\mathbf{A}_t = \mathbf{g}_\theta\!\big(V_t, T, Q\big),\\ 
\mathbf{A}_t[h]= a_{t+h},\; h=0,\ldots,H-1 .
\end{equation}
where $a_{t+h}$ denotes the $h$-step action and is trained with $\ell_1$ objective:
\begin{equation}
\mathcal{L}(\theta) = \!\big\|\mathbf{A}_t - \hat{\mathbf{A}}_t\big\|_1,
\end{equation}
yielding efficient one-shot continuous control under an $\ell_1$ objective. OpenVLA-OFT~\cite{kim2025fine} treats the zero embedding as a special query, using $Q=\mathbf{0}$ so that the VLM extracts information directly from $(V_t, T)$, and then decodes the action chunks directly from the last-layer query embedding with a simple action head. VLA-Adapter~\cite{wang2025vla}, in contrast, replaces this zero query with learnable queries $Q=\mathrm{Learnable}(N_q, d)$ and encodes vision and text conditions into the query latent by injecting $Q$ through per-layer \emph{bridge attention}~\cite{wang2025vla}, which modulates the $\tau$-th layer action latent $z^\tau$ via \textbf{S}elf-\textbf{A}ttention and \textbf{C}ross-\textbf{A}ttention:
 \\
\begin{equation}
\label{eq:action-expert}
\begin{array}{@{}l@{\quad}l@{}}
\shortstack[l]{\textbf{Action}\\\textbf{Expert}} &
\left\{
\begin{aligned}
z^{0} &= \mathbf{0},\\
C^{\tau} &= [\,\operatorname{SA}(z^\tau),\ \operatorname{CA}_{z^\tau}^{(t)}(\bar V^{\tau}),\ \operatorname{CA}_{z^\tau}(\bar Q^{\tau})\,],\\
z^{\tau+1} &= \operatorname{FFN}(C^{\tau}),\quad \tau=0,\ldots,L-1,\\
\mathbf{A}_t &= \operatorname{MLP}(z^{L})
\end{aligned}
\right.
\end{array}
\end{equation}
where $\tau$ indexes the layers of the VLM, $\operatorname{CA}_{z^\tau}^{(t)}(\bar V^{\tau})$ denotes cross-attention with $z^\tau$ as the query and the visual latents $\bar V^{\tau}$ from the VLM as keys/values, and the superscript $(t)$ indicates the $\tanh(g)$ gate applied to bound the output~\cite{zhang2023llama}.





\section{Modality Masking Mechanism (M3)}
\label{sec:method}





\paragraph{Design guidelines of M3.}

The core idea of M3 is to train the policy under controlled partial observability. Instead of always providing all multimodal tokens at every update, we stochastically hide subsets of modalities and action queries during training (Fig.~\ref{fig:overview}). The training objective remains the same regression objective, but the model is required to approximate the ground-truth even when some signals are missing. In our bimanual setting, we observe that naively dropping modalities at random can underperform a more structured approach, because careless masks may remove crucial global context, encourage diffuse attention patterns, or deprive the decoder of useful local signals. We therefore design M3 around three simple guidelines.

\noindent\textbf{\textit{Guideline 1. Preserve a stable spatial reference frame.}} We always keep the egocentric view visible to serve as a consistent spatial anchor across timesteps and masking realizations. Unlike wrist views—which capture task-relevant detail but are prone to occlusion, lighting variation, and distractor interference—the egocentric view provides a stable global frame for localizing targets and goals. Crucially, this is not a shortcut: masking the ego view instead is markedly worse ($37.0\%$ vs.\ $64.0\%$, Table~\ref{tab:ablation_vision_mask}), since the wrist views alone often cannot localize the goal (\eg, when the placement target lies outside both wrist fields of view). Instead, by keeping ego visible while randomly masking wrist views, we force the model to learn \emph{when and how} wrist views add value beyond what ego provides, rather than always fusing them indiscriminately. This encourages the policy to extract stable task geometry from the global view and only rely on local wrist views when they offer non-redundant, actionable evidence.

\noindent\textbf{\textit{Guideline 2. Mask dual-wrist views jointly, not independently.}} We always mask both wrist cameras together rather than masking them independently. The rationale is that the left and right wrist views are spatially correlated: both typically capture overlapping regions of the workspace, especially near the target object. If we mask them independently, the model can still perform spurious cross-view feature matching between the visible wrist view and the egocentric view (e.g., matching high-contrast edges or object saliency across views). By masking both wrist views jointly, we create a clean separation between global evidence (egocentric) and local evidence (wrist), forcing the policy to extract task geometry from the stable global frame rather than relying on which wrist view happens to have better lighting or fewer occlusions. This design reduces idle-arm interference: when wrist views are masked, the model cannot attend to distractors in the non-executing arm's camera, preventing the failure mode observed in Fig.~\ref{fig:heatmap} (top).

\noindent\textbf{\textit{Guideline 3. Apply query-subset masking to encourage specialization.}} For action queries, we never mask all queries simultaneously (which would make the task unsolvable), but instead stochastically drop a random subset at each training step. This follows the classical dropout principle~\cite{srivastava2014dropout}: when different subsets of queries must solve the same task under different masking patterns, they are discouraged from learning redundant representations (co-adaptation). Instead, queries specialize: some may focus on coarse waypoint planning, others on fine-grained contact alignment, and still others on temporal consistency across the action horizon. By forcing queries to remain useful even when their ``collaborators'' are absent, we encourage them to extract complementary, non-redundant information from the multimodal context. Empirically, the Q-group attention structure in Fig.~\ref{fig:mechanism_handover} (top) is more expressive under M3 than in the baseline, and we observe improved robustness when the policy encounters novel view configurations or partial occlusions at test time.

Empirically, our analysis (see Sec.~\ref{sec:analysis} and the \textbf{Appendix}) is consistent with these guidelines: (1) preserving the ego stream helps sustain global planning, (2) (L+R) arm-view masking reduces spurious cross-view interactions, and (3) query-subset masking helps the model use diverse semantic cues under dynamic masking.

\textit{M3 Strategy.}
We introduce a \emph{modality-level visibility}, which is integrated into the scaled dot-product attention through additive masks. This approach preserves all embeddings unchanged. Specifically, let \(V_t = [V_t^{\mathrm{ego}},\, V_t^{\mathrm{L}},\, V_t^{\mathrm{R}}]\) represent the visual token streams from the egocentric and two arm-mounted cameras. We propose to sample the mask from a Bernoulli distribution as \(u_v \sim \mathrm{Bernoulli}(1 - p_v)\) that masks both arm views \textbf{jointly}. Another mask is sampled from a Bernoulli distribution as \(u_\ell \sim \mathrm{Bernoulli}(1 - p_\ell)\) to mask the \textbf{entire} language modality, while ensuring the ego stream is always retained. For the action queries, we never fully drop any queries. Instead, we sample an element-wise vector \(u_q \in \{0, 1\}^{N_q}\) using \textit{i.i.d.} samples \(u_q(i) \sim \mathrm{Bernoulli}(1 - p_q)\) and enforce the constraint \(\|u_q\|_1 \geq 1\), ensuring at least one active query at all times.

\begin{figure*}[t!]
    \centering
    \includegraphics[width=\linewidth]{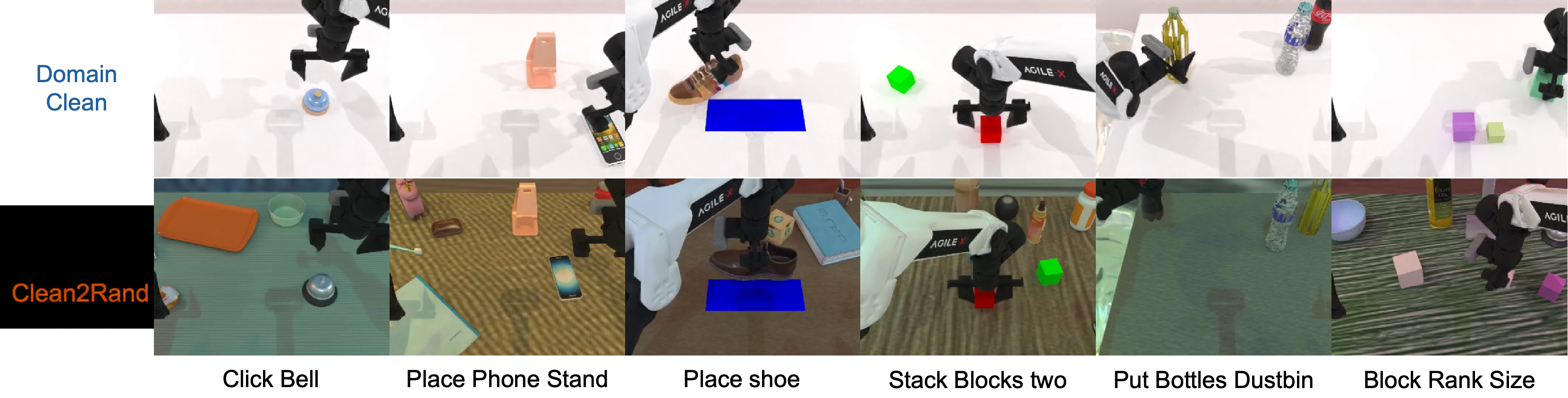}
    \caption{Example tasks from the RoboTwin 2.0~\cite{chen2025robotwin} benchmark, illustrating the visual distinction between the Clean (top row) and Clean2Rand (bottom row) evaluation settings. The six tasks shown are representative examples, with two selected from each of the short-, medium-, and long-horizon categories. Clean2Rand trains on clean demonstrations and evaluates on scenes with changed backgrounds, lighting, and distractor objects.}
    \Description{Six example RoboTwin 2.0 tasks are shown in two rows. The top row presents Clean evaluation scenes and the bottom row presents randomized scenes used for Clean2Rand evaluation across short-, medium-, and long-horizon tasks.}
    \label{fig:tasks}
\end{figure*}

We define the token index sets for all modalities, which partition the full sequence: \(\mathcal{S}_{\mathrm{ego}}\) (egocentric), \(\mathcal{S}_{\mathrm{wrist}}\) (wrist), \(\mathcal{S}_{\mathrm{lang}}\) (language), and \(\mathcal{S}_{q}\) (action queries). We then construct a single \emph{token visibility indicator} vector \(m\), where \(m_i=1\) if token \(i\) is visible and \(m_i=0\) otherwise. This vector is populated based on our sampled variables (letting \(i'\) be the relative index of \(i\) within \(\mathcal{S}_q\)):
\begin{equation}
m_i =
\begin{cases}
1, & i \in \mathcal{S}_{\mathrm{ego}}, \\
u_v, & i \in \mathcal{S}_{\mathrm{wrist}}, \\
u_\ell, & i \in \mathcal{S}_{\mathrm{lang}}, \\
u_q(i'), & i \in \mathcal{S}_{q}.
\end{cases}
\end{equation}
For self-attention, a masked token should neither ``see'' (row mask) nor ``be seen'' (column mask). We implement this with a single square additive mask \(\tilde{M}\), which permits attention only between two tokens if \textbf{both} are visible:
\begin{equation}
\tilde{M}_{xy} =
\begin{cases}
0, & m_x = 1 \text{ and } m_y = 1, \\
-\infty, & \text{otherwise}. 
\end{cases}
\end{equation}
Finally, the VLM using scaled dot-product attention with this unified modality mask is:
\begin{equation}
\operatorname{Attn} = \operatorname{softmax}\left( \frac{\mathcal{Q}\mathcal{K}^\top}{\sqrt{d}} + M_c + \tilde{M} \right) \mathcal{V},
\end{equation}
where \(M_c\) represents the causal mask and \(\tilde{M}\) is the unified modality-based mask. These masks are combined element-wise, enabling the model to train under controlled partial observability, with the goal of improving robustness under the input conditions.



\paragraph{Query Rescaling for Diverse Encoding.} 
In addition to the attention mask defined in $\tilde{M}$, we introduce a \textit{query rescaling} mechanism to promote diversity and robustness. This mechanism functions as \textit{dropout regularization along the query dimension}. Specifically, while the unified mask $\tilde{M}$ excludes masked queries from the attention computation, we further rescale the embeddings of the \emph{remaining} (visible) queries to maintain their expected magnitude, which can be formulated as
\begin{equation}
\tilde{\mathbf{q}}_i = \frac{1}{1-p_q} \cdot \mathbf{q}_i, \quad \text{if } u_q(i') = 1 \text{ (visible)}.
\end{equation}
This rescaling preserves the overall feature energy of the query embeddings. Consequently, the remaining queries may encode richer contextual information. Rather than relying on a fixed subset of queries, the model is encouraged to distribute task-relevant information across different query slots. Intuitively, each query may attend to distinct aspects of the visual-language context, since it cannot depend on a fixed partner query being consistently available. In our experiments, this dynamic adaptation is associated with improved robustness in downstream action prediction.

\begin{table*}[t!]
  \caption{Domain-clean performance across multi-horizon tasks on the \textbf{RoboTwin~2.0} simulation platform. 
  The best performance in each row is bolded, and the second-best is underlined.  $\Delta$ denotes the relative improvement of \textbf{M3} over the Adapter baseline. Models marked with $^{*}$ are pretrained on large-scale robot data, while our method achieves superior performance through direct fine-tuning.}
  \label{tab:performance_comparison}
  \centering
  \setlength{\tabcolsep}{12pt}
  \resizebox{\textwidth}{!}{%
  \begin{tabular}{@{}c|c|cc|cccc|c@{}}
    \toprule
    Category & Task Name & RDT$^{*}$ & $\pi_0^{*}$ & ACT & DP & Adapter & \cellcolor{gray!15}\textbf{M3} & $\Delta$ \\
    \midrule
    \multirow{3}{*}{Short Horizon} 
      & Click Bell & 80 & 44 & 58 & 54 & 84 & \cellcolor{gray!15}\textbf{97} & \textcolor{red!70!black}{+13} \\ 
      & Grab Roller & 74 & \underline{96} & 94 & \textbf{98} & 88 & \underline{96} & \textcolor{red!70!black}{+8} \\ 
      & Place Phone Stand & 15 & 35 & 2 & 13 & 10 & \cellcolor{gray!15}\textbf{55} & \textcolor{red!70!black}{+45} \\ 
    \midrule
    \multirow{4}{*}{Medium Horizon} 
      & Place Bread Basket & 10 & 17 & 6 & 14 & 11 & \cellcolor{gray!15}\textbf{22} & \textcolor{red!70!black}{+11} \\ 
      & Place A2B Right & 1 & 27 & 0 & 13 & 4 & \cellcolor{gray!15}\textbf{28} & \textcolor{red!70!black}{+24} \\ 
      & Place Shoe & 35 & 28 & 5 & 23 & 34 & \cellcolor{gray!15}\textbf{63} & \textcolor{red!70!black}{+29} \\
      & Stack Blocks Two & 21 & 42 & 25 & 7 & 78 & \cellcolor{gray!15}\textbf{83} & \textcolor{red!70!black}{+5} \\ 
    \midrule
    \multirow{3}{*}{Long Horizon} 
      & Handover Block & 45 & 45 & 42 & 10 & 27 & \cellcolor{gray!15}\textbf{74} & \textcolor{red!70!black}{+47} \\ 
      & Put Bottles Dustbin & 21 & 54 & 27 & 22 & 60 & \cellcolor{gray!15}\textbf{81} & \textcolor{red!70!black}{+21} \\ 
      & Block Rank Size & 0 & 7 & 0 & 1 & 14 & \cellcolor{gray!15}\textbf{28} & \textcolor{red!70!black}{+14} \\ 
    \midrule
  \multicolumn{2}{c|}{\textbf{Overall Avg}} 
    & 30.2 & 39.5 & 25.9 & 25.5 & 41.0 & \cellcolor{gray!15}\textbf{62.7} & \textcolor{red!70!black}{+21.7} \\
    \bottomrule
  \end{tabular}
  }
\end{table*}

\paragraph{M3 Objective.}
While adhering to the standard $\mathcal{L}_1$ loss used in baselines, our M3 training objective is applied to predictions conditioned on M3-masked inputs. Formally, let $V_t^{M3}$, $T_t^{M3}$, and $Q_t^{M3}$ denote the visible visual tokens, language tokens, and action queries after applying M3. The training objective remains the same $\mathcal{L}_1$ regression loss:
\begin{equation}
    \mathcal{L}_{M3}(\theta) = ||\mathbf{g}_{\theta}(V_t^{M3}, T_t^{M3}, Q_t^{M3})-\mathbf{\hat{A}}_t||_1,
\end{equation}
where $g_{\theta}$ denotes the policy. As illustrated in Fig.~\ref{fig:overview}, we follow the Adapter-style action expert, which decodes actions from visual latents and query latents that already carry the language signal. Training $g_{\theta}$ to approximate $\hat{A}_t$ under deliberate partial observability may encourage the query-based VLA model to learn more robust representations. In practice, this setup may help the model extract diverse, complementary information from these ``dynamically observable latents,'' potentially improving generalization.

\section{Experiments}
\label{sec:experiments}

\subsection{Setup}
\label{sec:setup}
\paragraph{Benchmark.} We adopt \textbf{RoboTwin~2.0}~\cite{chen2025robotwin} as our primary simulation benchmark. The platform offers \emph{50} dual-arm manipulation tasks with domain randomization over distractors, backgrounds, lighting, table heights, and language instructions. Following~\cite{li2025simplevla}, we select \emph{10} tasks grouped into \emph{short-, medium-, and long-horizon} categories by mean step count (see Fig.~\ref{fig:tasks}). Each task is trained on 50 clean demonstrations and evaluated on 100 held-out scenarios in two settings: Clean uses held-out clean scenes, while Clean2Rand evaluates the same clean-trained policy on held-out randomized scenes. For simulation, we follow the official RoboTwin benchmark evaluation protocol and report the corresponding single-run results for direct comparison with prior work.
\textbf{Real-world protocol.} We further construct three long-horizon bimanual tasks on the \textit{Agilex Cobot} platform: bottle disposal with inter-arm handover, bowl stacking and shelving, and vegetable plating followed by plate centering. Each task exceeds \(800\) control steps and involves multi-stage grasping, handover, and coordinated transport. We collect \textbf{50} demonstrations per task and evaluate under two settings: \emph{clean}, which uses the nominal layout with only task-relevant objects, and \emph{OOD}, which introduces novel distractor objects near the targets following recent generalization protocols~\cite{li2024cogact,zitkovich2023rt}. For each task and each model, we repeat the real-world evaluation for three rounds: each round contains 16 clean trials and 8 OOD trials. For fair repeated measurement, the corresponding trial in each round uses the same scene layout, including the distractor arrangement in the OOD setting. We report averaged full-task success rates across the three rounds, corresponding to 48 clean and 24 OOD trials per task in total. Full details appear in Fig.~\ref{fig:real_world_platform} and \textbf{Appendix}.
\paragraph{Implementation details.}
Our VLM backbone is the Prismatic VLM~\cite{karamcheti2024prismatic}, built upon the Qwen2.5-$0.5$B~\cite{Yang2024Qwen25TR} language model.
A key component is its hybrid vision encoder, which combines DINOv2~\cite{dinov2} and SigLIP~\cite{zhai2023sigmoid}, an architectural choice consistent with OpenVLA~\cite{kim2024openvla}. Furthermore, our \textit{action expert} module is designed in alignment with the Adapter~\cite{wang2025vla} approach, where, consistent with the baseline, proprioceptive inputs are utilized as default keys and values. Unless otherwise specified, all models use the \textit{AdamW} optimizer, an initial learning rate of $2e-4$, and are trained for~$10\,\mathrm{k}$ steps with a MultiStep decay of~$\times 0.1$ at~$5\,\mathrm{k}$ steps.
VLA-Adapter follows the official \textit{pro} configuration as a strong baseline. Our method, M3, preserves the backbone architecture and introduces only training-time modality masking, and the LoRA rank is fixed to~$64$. We use identical hyperparameters across short/medium/long horizons in all experiments. Additional implementation details are provided in the \textbf{Appendix}.

\subsection{Main Results}
\label{sec:main_results}

\paragraph{Domain-clean results.}
On RoboTwin~2.0~\cite{chen2025robotwin} (Table~\ref{tab:performance_comparison}), M3 improves the overall success rate from \(41.0\%\) to \(62.7\%\), outperforming the Adapter baseline as well as the compared pretrained policies in this evaluation.%
~The gains are especially pronounced on long-horizon tasks, where the average rises from \(33.7\%\) to \(61.0\%\); on \textit{Handover Block} alone, success improves from \(27\%\) to \(74\%\).%
~As shown in the bottom panel of Fig.~\ref{fig:mechanism_handover}, M3 also exhibits more stable execution on this representative long-horizon task.%
~These results indicate that structured training-time masking can substantially improve robustness for this query-based bimanual VLA setting.

\paragraph{Clean2Rand results.}
In Clean2Rand evaluation (Table~\ref{tab:clean2rand_performance_comparison}), M3 achieves \(15.1\%\) overall success, a \(+11.4\%\) gain over the Adapter baseline and the highest among all compared methods.%
~The largest per-task gains appear on \textit{Put Bottles Dustbin} \((+38\%\)) and \textit{Grab Roller} \((+26\%\)).%
~The sole exception is \textit{Block Rank Size}, which stays near zero for every method. This task demands fine-grained relational size reasoning~\cite{intelligence2025pi_} and already exhibits residual diffuse attention in the Clean setting (Fig.~\ref{fig:heatmap}), and the distractors in Clean2Rand further amplify this cross-view interference.%
~Overall, these results indicate that \emph{training-only} masking can meaningfully improve domain-shift robustness without large-scale robot data.

\begin{table}[tb!]
\caption{\textbf{Clean2Rand performance} across multi-horizon tasks on the \textbf{RoboTwin~2.0} simulation platform. All task-specific policies are trained on 50 clean demonstrations and evaluated on 100 held-out randomized scenes. Relative to the fine-tuned baseline, \textbf{M3} improves average success by \textbf{11.4\%}. Unlike VLA policies such as $\pi_0$~\cite{black2024pi_0} and RDT-1B~\cite{liu2024rdt}, which rely on massive cross-embodiment, internet-scale or multi-robot pretraining to reduce the domain gap, \textbf{M3} uses only clean, task-specific training data and achieves comparable or better generalization.}
\label{tab:clean2rand_performance_comparison}
\centering
\setlength{\tabcolsep}{12pt}
\resizebox{\textwidth}{!}{%
\begin{tabular}{@{}c|c|cc|cccc|c@{}}
  \toprule
  Category & Task Name & RDT$^{*}$ & $\pi_0^{*}$ & ACT & DP & Adapter & \cellcolor{gray!15}\textbf{M3} & $\Delta$ \\
  \midrule
  \multirow{3}{*}{Short Horizon}
    & Click Bell & 9 & 3 & 3 & 0 & 7 & \cellcolor{gray!15}16 & \textcolor{red!70!black}{+9} \\
    & Grab Roller & 43 & 80 & 25 & 0 & 28 & \cellcolor{gray!15}54 & \textcolor{red!70!black}{+26} \\
    & Place Phone Stand & 6 & 7 & 0 & 0 & 0 & \cellcolor{gray!15}12 & \textcolor{red!70!black}{+12} \\
  \midrule
  \multirow{4}{*}{Medium Horizon}
    & Place Bread Basket & 2 & 4 & 0 & 0 & 0 & \cellcolor{gray!15}6 & \textcolor{red!70!black}{+6} \\
    & Place A2B Right & 1 & 6 & 0 & 0 & 0 & \cellcolor{gray!15}2 & \textcolor{red!70!black}{+2} \\
    & Place Shoe & 7 & 6 & 0 & 0 & 0 & \cellcolor{gray!15}11 & \textcolor{red!70!black}{+11} \\
    & Stack Blocks Two & 2 & 1 & 0 & 0 & 0 & \cellcolor{gray!15}4 & \textcolor{red!70!black}{+4} \\
  \midrule
  \multirow{3}{*}{Long Horizon}
    & Handover Block & 14 & 8 & 0 & 0 & 0 & \cellcolor{gray!15}6 & \textcolor{red!70!black}{+6} \\
    & Put Bottles Dustbin & 4 & 13 & 1 & 0 & 2 & \cellcolor{gray!15}40 & \textcolor{red!70!black}{+38} \\
    & Block Rank Size & 0 & 1 & 0 & 0 & 0 & \cellcolor{gray!15}0 & \textcolor{red!70!black}{0} \\
  \midrule
  \multicolumn{2}{c|}{\textbf{Overall Avg}}
    & 8.8 & 12.9 & 2.9 & 0.0 & 3.7 & \cellcolor{gray!15}\textbf{15.1} & \textcolor{red!70!black}{+\textbf{11.4}} \\
  \bottomrule
\end{tabular}}%
\end{table}

\begin{figure}[t]
  \centering
  \vspace{-1em}
  \includegraphics[width=0.80\columnwidth]{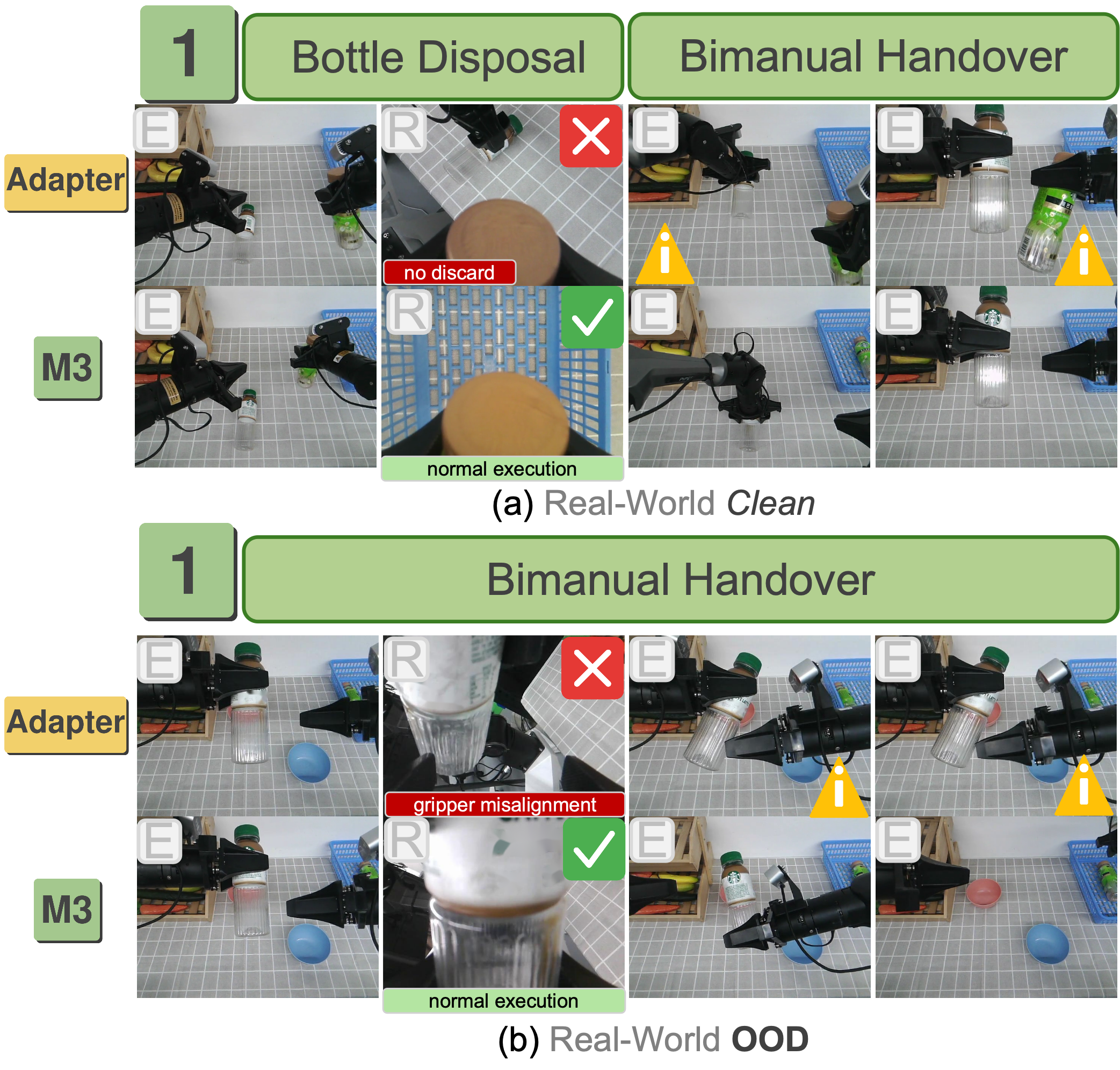}
  \caption{\textbf{Qualitative comparison of real-world execution under clean and OOD conditions.} The second column provides a close-up from the right-arm wrist camera. (a)~In the clean setting, the Adapter baseline fails to release the bottle into the bin, whereas M3 completes the discard smoothly. (b)~Under OOD clutter, the baseline suffers from gripper misalignment during handover, while M3 maintains stable alignment throughout the sequence.}
  \Description{Two real-world rollout panels compare the Adapter baseline and M3 in clean and out-of-distribution settings. In the clean setting, the baseline fails to discard the bottle into the bin. In the OOD setting, the baseline exhibits gripper misalignment during handover. M3 completes the corresponding sequences more smoothly.}
  \label{fig:realworld_qualitative}
\end{figure}

\paragraph{Real-world results.}
Fig.~\ref{fig:realworld_qualitative} compares representative rollouts under both the \emph{clean} and \emph{OOD} settings. In the clean setting, the Adapter baseline fails to discard the bottle into the bin, whereas M3 completes the task sequence successfully in this example rollout. Under the OOD setting, where novel distractor objects are placed near the targets, the baseline exhibits gripper misalignment during the inter-arm handover, while M3 appears more stable throughout the episode. More cases are shown in the \textbf{Appendix}. Quantitatively (Fig.~\ref{fig:real_world_platform}), across three repeated evaluation rounds totalling \(144\) clean and \(72\) OOD trials, M3 raises the averaged full-task success rate from \(44.4\%\) to \(69.4\%\) under clean conditions and from \(12.5\%\) to \(61.1\%\) under OOD conditions. These numbers suggest that novel distractors unseen during training substantially reduce the baseline's full-task success, whereas the M3-trained policy appears less affected under the same OOD protocol.

\FloatBarrier

\section{Analysis}
\label{sec:analysis}

\begin{wrapfigure}{r}{0.50\textwidth}
  \centering
  \captionsetup{font=small,skip=4pt}
  \includegraphics[width=\linewidth]{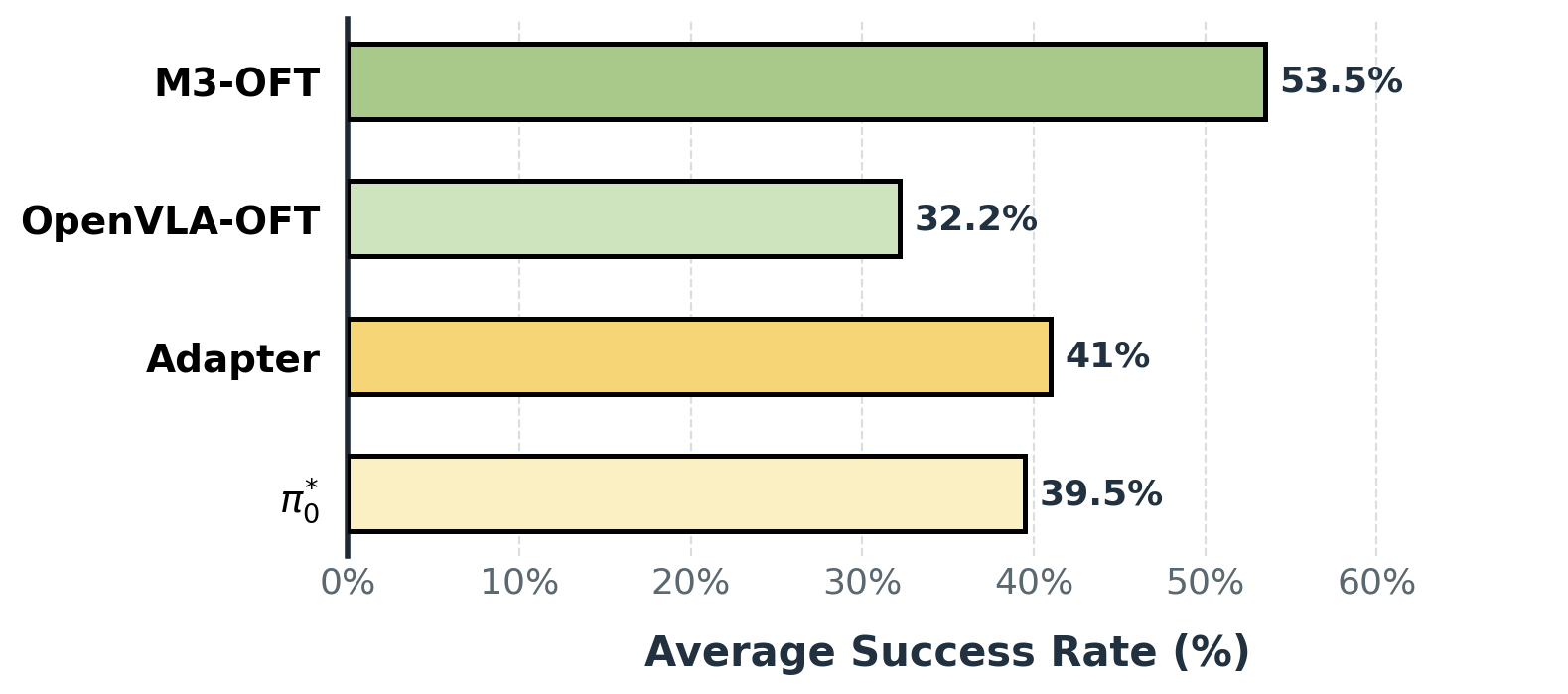}
  \caption{\textbf{Cross-backbone transfer.} Under the domain-clean RoboTwin protocol, M3-OFT raises average success from $32.2\%$ to $53.5\%$ and exceeds the evaluated baselines.}
  \Description{Horizontal bar chart of average success rate under the domain-clean RoboTwin protocol for M3-OFT at 53.5 percent, OpenVLA-OFT at 32.2 percent, Adapter at 41.0 percent, and pi-zero star at 39.5 percent.}
  \label{fig:cross_backbone_evidence}
\end{wrapfigure}

\paragraph{Cross-backbone transfer.} To probe whether M3 may transfer beyond the Adapter backbone, we apply it to OpenVLA-OFT~\cite{kim2025fine}, a structurally distinct query-based VLA with parallel decoding and bidirectional attention. As shown in Fig.~\ref{fig:cross_backbone_evidence}, M3 improves the overall average success rate of OpenVLA-OFT from $32.2\%$ to $53.5\%$ under the domain-clean RoboTwin protocol, with gains across short-, medium-, and long-horizon tasks. Taken together, these results suggest that the benefits of the proposed structured partial-observability training strategy may extend beyond the primary backbone studied in the main paper. Further per-task breakdowns are provided in the \textbf{Appendix}.

\begin{figure*}[t]
  \centering
  \includegraphics[width=\textwidth]{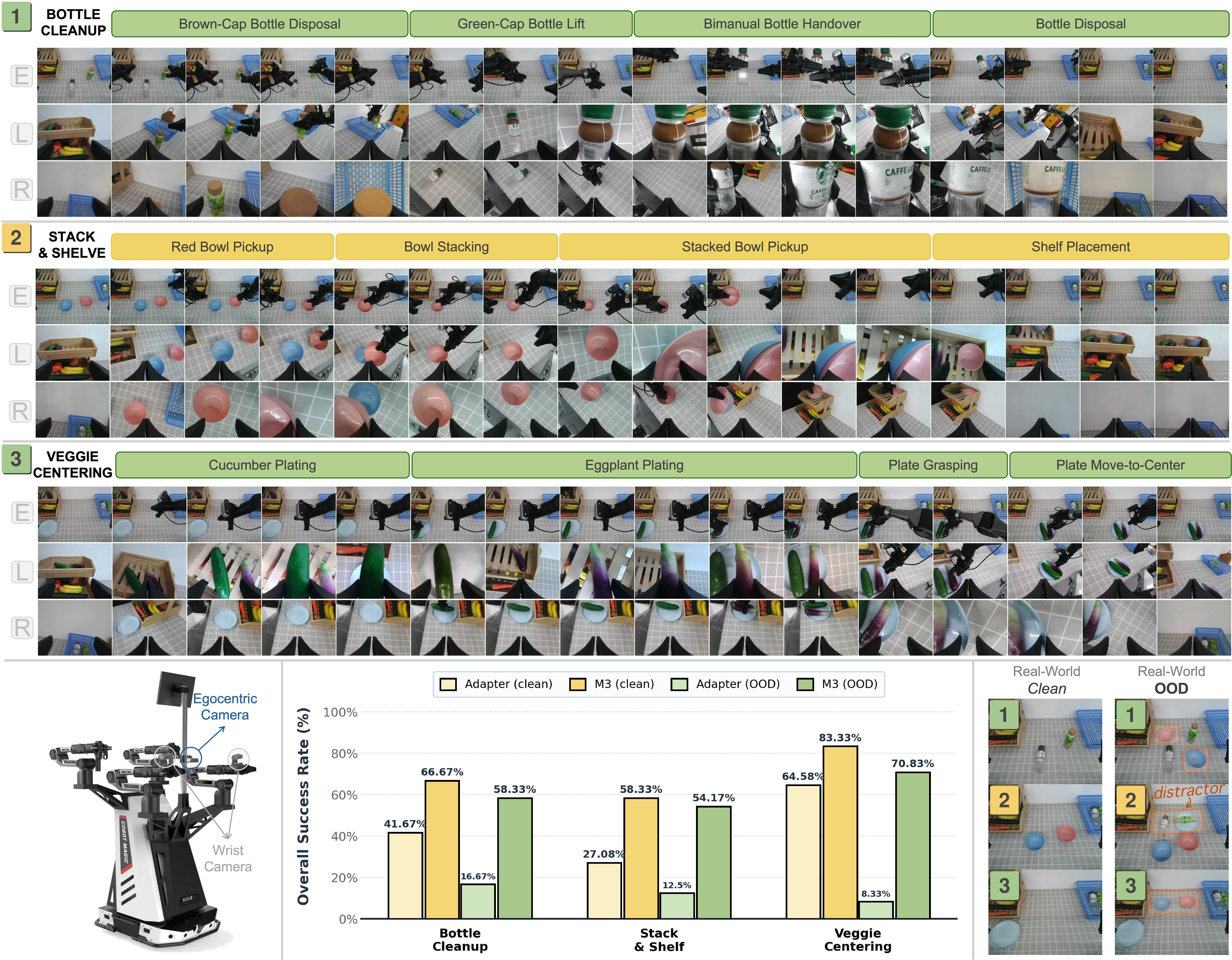}
  \caption{Overview of real-world evaluation. \textbf{Top:} three representative tasks successfully executed by M3: (1)~Bottle Cleanup, (2)~Stack Shelf, and (3)~Veggie Centering. \textbf{Bottom-left:} our bimanual robot platform. \textbf{Bottom-center:} full-task success rates under clean and OOD settings, where M3 shows consistent improvement over the Adapter baseline. \textbf{Bottom-right:} illustration of the two evaluation settings (clean vs.\ OOD with novel distractor objects). The baseline degrades more sharply under OOD clutter, suggesting that M3 appears to be more robust to unseen distractors. Detailed task definitions, protocols, additional demonstrations, and tabulated results are given in Sec.~\ref{sec:setup} and the \textbf{Appendix}.}
  \Description{A composite real-world evaluation figure shows representative image sequences for Bottle Cleanup, Stack Shelf, and Veggie Centering, a photo of the dual-arm robot platform, grouped bar charts of clean and OOD success rates for Adapter and M3, and small thumbnails illustrating the clean and distractor-rich evaluation settings.}
  \label{fig:real_world_platform}
\end{figure*}
\FloatBarrier

\begin{figure}[t!]
    \centering
    \captionsetup{font=small,skip=4pt}
    \includegraphics[width=0.78\linewidth]{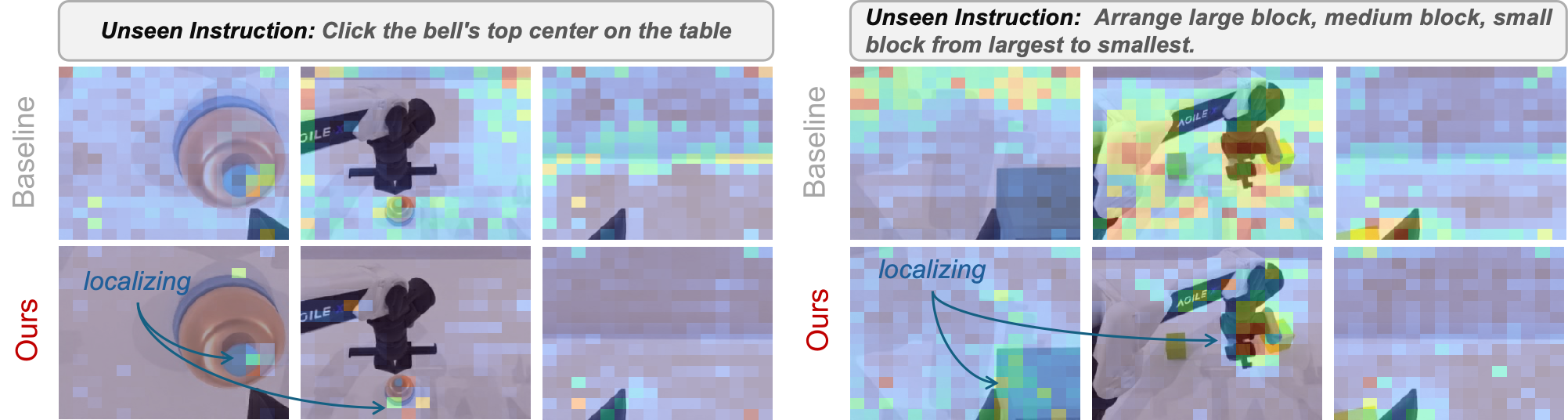}
    \caption{\textbf{Spatial attention under viewpoint variation.} On \textit{Click Bell} and \textit{Block Rank Size}, the Adapter baseline spreads attention over distractors, the gripper, and background regions, whereas M3 more consistently concentrates on the target and contact region. Attention remains partly diffuse on the harder \textit{Block Rank Size}.}
    \Description{A grid of attention heatmaps compares the Adapter baseline and M3 on Click Bell and Block Rank Size using the ego and wrist cameras. Baseline attention is diffuse and often drifts to distractors, while M3 is more concentrated on the target object and hand-object contact region.}
    \label{fig:heatmap}
\end{figure}

\paragraph{Visualization of M3.} Fig.~\ref{fig:heatmap} compares spatial attention maps of the baseline and M3 on RoboTwin~2.0 tasks. We consistently observe that the baseline allocates more attention to distractors, the robot gripper, and background regions, whereas M3 tends to produce more compact attention around the object and contact region. The token-level attention-score analysis in the top panel of Fig.~\ref{fig:mechanism_handover} complements these spatial maps by visualizing interactions among the egocentric, wrist, language, and action-query groups. Together with complementary statistics in the \textbf{Appendix}, these patterns are consistent with our hypothesis that training under structured partial observability encourages more task-focused evidence selection; concurrent work likewise uses attention analysis to study VLA failures~\cite{huang2025otter,song2025reconvla}.

\begin{figure}[H]
  \centering
  \includegraphics[width=0.82\linewidth]{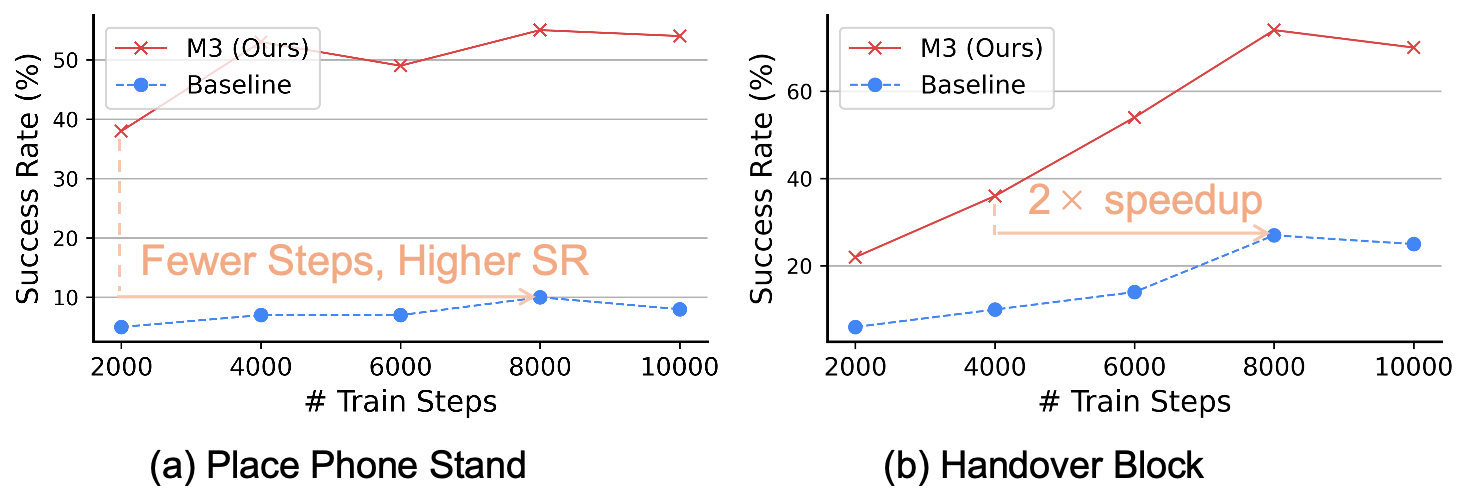}
  \caption{Success rate vs.\ training steps on \textit{Place Phone Stand} (short-horizon) and \textit{Handover Block} (long-horizon). Under the same compute budget, M3 tends to reach higher success rates earlier and maintains a consistent margin over the Adapter baseline throughout training.}
  \Description{Two line charts plot success rate against training steps for Place Phone Stand and Handover Block. In both tasks, the M3 curve rises faster than the Adapter baseline and stays at a higher success level later in training.}
  \label{fig:train_efficiency}
\end{figure}

\paragraph{Training efficiency.} We define training efficiency as achieving a target success rate with fewer optimization steps while preserving or improving the final plateau. Fig.~\ref{fig:train_efficiency} compares learning curves on a short-horizon task (Place Phone Stand) and a long-horizon task (Handover Block) under identical settings. Across both settings, M3 rises more steeply and saturates at a higher level than the Adapter baseline. It reaches strong performance earlier and sustains a stable margin thereafter, indicating better sample efficiency and more reliable convergence. This trend suggests that M3 may provide a more favorable training signal in these settings.

\paragraph{Ablation of M3.}
At the view level, Table~\ref{tab:targeted_masking_ablations}(a) shows that preserving the egocentric (ego) view while jointly masking both wrist views outperforms masking the ego view or only one wrist view. This supports the role of the ego view as a global anchor and suggests that paired wrist-view masking is preferable to independent single-wrist masking. Table~\ref{tab:targeted_masking_ablations}(b) further shows that a light query-mask ratio works best, whereas overly aggressive query masking reduces success. Table~\ref{tab:ablation_m3}(a) then decomposes M3 into vision, language, and query masking components. Single-component masking remains limited (29.0\%--34.0\% Avg.~SR), and adding language to vision gives only 36.7\%; in contrast, combining vision and query masking raises Avg.~SR to 60.3\%, while the full V+L+Q configuration reaches 64.0\%. Table~\ref{tab:ablation_m3}(b) compares M3 with generic regularization baselines: token dropout (31.8\%), modality dropout (24.1\%), visual augmentation (22.3\%), and region augmentation (23.0\%) all stay far below full M3. Overall, Table~\ref{tab:ablation_m3} indicates that the main gain comes from coupling structured vision masking with query masking, and that generic dropout or augmentation is not sufficient for this bimanual multi-view setting.

\begin{table}[H]
    \centering
    \caption{\textbf{Targeted masking ablations.} \textbf{(a)} Vision view-level masking while non-view masking components follow the full-M3 setting. Rows compare masking the egocentric view, one wrist view (L or R), and both wrist views. \textbf{(b)} Query-mask ratio, including the unmasked Adapter baseline (0.0). Masking both wrist views and using a light query-mask ratio (0.1) achieve the highest average success rates in their respective comparisons.}
    \label{tab:targeted_masking_ablations}
    \begin{subtable}[t]{0.49\textwidth}
        \centering
        \caption{Vision view-level masking.}
        \label{tab:ablation_vision_mask}
        \resizebox{\linewidth}{!}{
        \begin{tabular}{c|cccc}
            \toprule
            \textbf{M3 variant} & Place Phone Stand & Place Shoe & Handover Block & Avg \\
            \midrule
            Mask ego & 20 & 49 & 42 & 37.0 \\
            Mask one wrist & 22 & 39 & 37 & 32.7 \\
            \textbf{Mask both wrists} & \textbf{55} & \textbf{63} & \textbf{74} & \textbf{64.0} \\
            \bottomrule
        \end{tabular}
        }
    \end{subtable}\hfill
    \begin{subtable}[t]{0.49\textwidth}
        \centering
        \caption{Query-mask ratio.}
        \label{tab:query_mask_ratio}
        \resizebox{\linewidth}{!}{
        \begin{tabular}{c|cccc}
            \toprule
            Query Mask Ratio & Place Phone Stand & Place Shoe & Handover Block & Avg \\
            \midrule
            0.0 (Adapter) & 10 & 34 & 27 & 23.7 \\
            \midrule
            \textbf{0.1} & \textbf{55} & \textbf{63} & \textbf{74} & \textbf{64.0} \\
            0.3 & 50 & 59 & 59 & 56.0 \\
            0.5 & 49 & 58 & 63 & 56.7 \\
            0.7 & 52 & 55 & 55 & 54.0 \\
            0.9 & 41 & 40 & 50 & 43.7 \\
            \bottomrule
        \end{tabular}
        }
    \end{subtable}
\end{table}

\begin{table}[H]
    \centering
    \caption{\textbf{Ablation studies.} \textbf{(a)} Component ablation: V, L, and Q denote vision, language, and query masking. \textbf{(b)} Comparison with dropout and augmentation baselines. Avg.~SR is the mean success rate over \textit{Place Phone Stand}, \textit{Place Shoe}, and \textit{Handover Block}.}
    \label{tab:ablation_m3}
    \vspace{0.5em}
    \begin{subtable}[t]{0.48\linewidth}
        \centering
        \caption{Component ablation.}
        \small
        \setlength{\tabcolsep}{2pt}
        \renewcommand{\arraystretch}{0.95}
        \begin{tabular}{@{}llc@{}}
            \toprule
            \textbf{Method} & \textbf{Mask} & \textbf{Avg.~SR} \\
            \midrule
            Adapter & none & 23.7 \\
            Language-only & L & 29.0 \\
            Vision-only & V & 34.0 \\
            Query-only & Q & 33.3 \\
            Vision+language & V+L & 36.7 \\
            Vision+query & V+Q & 60.3 \\
            \rowcolor{gray!15}\textit{M3 (full)} & V+L+Q & \textbf{64.0} \\
            \bottomrule
        \end{tabular}
    \end{subtable}
    \hfill
    \begin{subtable}[t]{0.48\linewidth}
        \centering
        \caption{Dropout and augmentation baselines.}
        \small
        \setlength{\tabcolsep}{2pt}
        \renewcommand{\arraystretch}{0.95}
        \begin{tabular}{@{}lc@{}}
            \toprule
            \textbf{Method} & \textbf{Avg.~SR} \\
            \midrule
            Adapter & 23.7 \\
            Token dropout & 31.8 \\
            Modality dropout & 24.1 \\
            Visual aug. & 22.3 \\
            Region aug. & 23.0 \\
            Mask one wrist & 32.7 \\
            \rowcolor{gray!15}\textit{M3 (full)} & \textbf{64.0} \\
            \bottomrule
        \end{tabular}
    \end{subtable}
\end{table}

\FloatBarrier

\section{Related Works}
\label{sec:related_works}
\paragraph{Vision-Language-Action Models.} VLAs~\cite{zitkovich2023rt,kim2024openvla,kim2025fine,black2024pi_0,liu2024rdt,zhong2025survey,shao2025large,bjorck2025gr00t,li2026light} have greatly advanced general-purpose robotic policies by leveraging large-scale pretrained vision-language models (VLMs)~\cite{beyer2024paligemma,karamcheti2024prismatic,bai2025qwen2} to connect perception with action. Current VLAs generally fall into three modeling categories. (1) Autoregressive models such as RT-2~\cite{zitkovich2023rt} and OpenVLA~\cite{kim2024openvla} discretize continuous actions into language tokens and predict them sequentially. CoT-VLA~\cite{zhao2025cot} further introduces Chain-of-Thought reasoning to imagine intermediate visual states before generating actions, separating visual reasoning from execution. However, such discretization can lead to quantization errors that reduce action precision in fine-grained manipulation. (2) Diffusion-based models operate directly in continuous action space. Diffusion Policy~\cite{chi2025diffusion} learns high-quality grasp trajectories through iterative denoising, while $\pi_0$~\cite{black2024pi_0} enhances sampling efficiency via a Mixture-of-Transformers~\cite{liang2024mixture} and flow matching~\cite{lipman2022flow}. Although effective, these approaches~\cite{kim2024openvla,black2024pi_0} often depend on extensive pretraining. (3) Query-based models decode actions in a single forward pass and regression objective. ACT~\cite{zhao2023learning} adopts a DETR-style architecture with temporal ensembling for stable predictions, and OpenVLA-OFT~\cite{kim2025fine} extends it with parallel decoding for efficient adaptation. VLA-Adapter~\cite{wang2025vla} further scales this design to a compact 0.5B-parameter model, introducing action queries and bridge-attention experts, achieving competitive single-arm performance without large-scale retraining. Despite these advances, query-based VLAs remain underexplored in more complex bimanual settings~\cite{chen2025robotwin}. To address this gap, we introduce a lightweight modality-masking mechanism that requires no architectural changes and incurs zero additional compute or parameters, and empirically improves policy generalization across diverse bimanual manipulation tasks.

\paragraph{Learning-based Bimanual Manipulation.} Leveraging structured reinforcement learning and coordination priors, recent bimanual manipulation works have achieved robust sim-to-real transfer across various dual-arm skills~\cite{fan2024learning,cui2024task}. Concurrently, multimodal perception, including visuo-tactile fusion and wrist-force sensing, has improved contact state estimation, robustness, and data efficiency~\cite{huang20243d,lin2023bi,stepputtis2022system}. Moreover, directly learning implicit action signals from human videos bridges the embodiment gap and lowers demonstration costs~\cite{bahety2024screwmimic,zhou2025you,zhou2025binomap}. Within the most widely studied imitation-learning track, ACT and diffusion policies on low-cost and mobile whole-body ALOHA systems demonstrate strong real-world coordination under modest supervision~\cite{zhao2023learning,chi2025diffusion,fu2024mobile,zhao2024aloha}; foundation-scale diffusion (RDT-1B)~\cite{liu2024rdt} and efficient 3D flow-matching policies~\cite{gkanatsios20253d} unify action spaces and set new state of the art across robots and tasks. However, diffusion models can incur nontrivial training and inference cost; our work instead explores a query-based foundation policy that directly decodes actions in a single step, aiming to retain efficiency while matching or surpassing pretrained models.

\paragraph{Dropout and training-time regularization.} Dropout~\cite{srivastava2014dropout} randomly deactivates neurons during training to prevent co-adaptation. This principle has been extended to layer dropout~\cite{huang2016deep}, attention dropout~\cite{vaswani2017attention}, and modality dropout for multimodal fusion~\cite{neverova2016moddrop,liu2017sensor}. Our ablations (Table~\ref{tab:ablation_m3}) show that generic token dropout and modality dropout reach only 31.8\% and 24.1\% average success, respectively, whereas the structured M3 configuration reaches 64.0\%.

\section{Limitations}
\label{sec:limitations}

While M3 shows consistent gains across RoboTwin 2.0 and three real-world tasks, broader evaluation on additional robot platforms, task categories, wider VLA architectures, and richer input modalities such as 3D geometric information, together with a deeper analysis of how structured masking influences multimodal fusion and diffusion-based action decoding, remains future work.

\section{Conclusion}
\label{sec:conclusion}
We study robustness issues in query-based VLA models for bimanual manipulation and introduce M3, a simple training-only modality masking strategy. Across RoboTwin~2.0 and three long-horizon real-world tasks, M3 consistently improves over the Adapter baseline while requiring no inference-time architectural changes. These results suggest that structured training-time masking is a practical way to strengthen query-based bimanual VLA policies.

\bibliography{references}

@article{chen2025robotwin,
  title={Robotwin 2.0: A scalable data generator and benchmark with strong domain randomization for robust bimanual robotic manipulation},
  author={Chen, Tianxing and Chen, Zanxin and Chen, Baijun and Cai, Zijian and Liu, Yibin and Li, Zixuan and Liang, Qiwei and Lin, Xianliang and Ge, Yiheng and Gu, Zhenyu and others},
  journal={arXiv preprint arXiv:2506.18088},
  year={2025}
}

@article{kim2025fine,
  title={Fine-tuning vision-language-action models: Optimizing speed and success},
  author={Kim, Moo Jin and Finn, Chelsea and Liang, Percy},
  journal={arXiv preprint arXiv:2502.19645},
  year={2025}
}

@article{liu2024rdt,
  title={Rdt-1b: a diffusion foundation model for bimanual manipulation},
  author={Liu, Songming and Wu, Lingxuan and Li, Bangguo and Tan, Hengkai and Chen, Huayu and Wang, Zhengyi and Xu, Ke and Su, Hang and Zhu, Jun},
  journal={arXiv preprint arXiv:2410.07864},
  year={2024}
}

@article{black2024pi_0,
  title={$\pi_0$: A Vision-Language-Action Flow Model for General Robot Control},
  author={Black, Kevin and Brown, Noah and Driess, Danny and Esmail, Adnan and Equi, Michael and Finn, Chelsea and Fusai, Niccolo and Groom, Lachy and Hausman, Karol and Ichter, Brian and others},
  journal={arXiv preprint arXiv:2410.24164},
  year={2024}
}

@article{li2024cogact,
  title={Cogact: A foundational vision-language-action model for synergizing cognition and action in robotic manipulation},
  author={Li, Qixiu and Liang, Yaobo and Wang, Zeyu and Luo, Lin and Chen, Xi and Liao, Mozheng and Wei, Fangyun and Deng, Yu and Xu, Sicheng and Zhang, Yizhong and others},
  journal={arXiv preprint arXiv:2411.19650},
  year={2024}
}

@article{zhao2023learning,
  title={Learning fine-grained bimanual manipulation with low-cost hardware},
  author={Zhao, Tony Z and Kumar, Vikash and Levine, Sergey and Finn, Chelsea},
  journal={arXiv preprint arXiv:2304.13705},
  year={2023}
}

@article{chi2025diffusion,
  title={Diffusion policy: Visuomotor policy learning via action diffusion},
  author={Chi, Cheng and Xu, Zhenjia and Feng, Siyuan and Cousineau, Eric and Du, Yilun and Burchfiel, Benjamin and Tedrake, Russ and Song, Shuran},
  journal={The International Journal of Robotics Research},
  volume={44},
  number={10-11},
  pages={1684--1704},
  year={2025},
  publisher={Sage Publications Sage UK: London, England}
}

@inproceedings{zitkovich2023rt,
  title={Rt-2: Vision-language-action models transfer web knowledge to robotic control},
  author={Zitkovich, Brianna and Yu, Tianhe and Xu, Sichun and Xu, Peng and Xiao, Ted and Xia, Fei and Wu, Jialin and Wohlhart, Paul and Welker, Stefan and Wahid, Ayzaan and others},
  booktitle={Conference on Robot Learning},
  pages={2165--2183},
  year={2023},
  organization={PMLR}
}

@article{zhong2025survey,
  title={A Survey on Vision-Language-Action Models: An Action Tokenization Perspective},
  author={Zhong, Yifan and Bai, Fengshuo and Cai, Shaofei and Huang, Xuchuan and Chen, Zhang and Zhang, Xiaowei and Wang, Yuanfei and Guo, Shaoyang and Guan, Tianrui and Lui, Ka Nam and others},
  journal={arXiv preprint arXiv:2507.01925},
  year={2025}
}

@article{shao2025large,
  title={Large vlm-based vision-language-action models for robotic manipulation: A survey},
  author={Shao, Rui and Li, Wei and Zhang, Lingsen and Zhang, Renshan and Liu, Zhiyang and Chen, Ran and Nie, Liqiang},
  journal={arXiv preprint arXiv:2508.13073},
  year={2025}
}

@article{kim2024openvla,
  title={Openvla: An open-source vision-language-action model},
  author={Kim, Moo Jin and Pertsch, Karl and Karamcheti, Siddharth and Xiao, Ted and Balakrishna, Ashwin and Nair, Suraj and Rafailov, Rafael and Foster, Ethan and Lam, Grace and Sanketi, Pannag and others},
  journal={arXiv preprint arXiv:2406.09246},
  year={2024}
}

@inproceedings{zhao2025cot,
  title={Cot-vla: Visual chain-of-thought reasoning for vision-language-action models},
  author={Zhao, Qingqing and Lu, Yao and Kim, Moo Jin and Fu, Zipeng and Zhang, Zhuoyang and Wu, Yecheng and Li, Zhaoshuo and Ma, Qianli and Han, Song and Finn, Chelsea and others},
  booktitle={Proceedings of the Computer Vision and Pattern Recognition Conference},
  pages={1702--1713},
  year={2025}
}

@article{liang2024mixture,
  title={Mixture-of-transformers: A sparse and scalable architecture for multi-modal foundation models},
  author={Liang, Weixin and Yu, Lili and Luo, Liang and Iyer, Srinivasan and Dong, Ning and Zhou, Chunting and Ghosh, Gargi and Lewis, Mike and Yih, Wen-tau and Zettlemoyer, Luke and others},
  journal={arXiv preprint arXiv:2411.04996},
  year={2024}
}

@article{wang2025vla,
  title={Vla-adapter: An effective paradigm for tiny-scale vision-language-action model},
  author={Wang, Yihao and Ding, Pengxiang and Li, Lingxiao and Cui, Can and Ge, Zirui and Tong, Xinyang and Song, Wenxuan and Zhao, Han and Zhao, Wei and Hou, Pengxu and others},
  journal={arXiv preprint arXiv:2509.09372},
  year={2025}
}

@article{lipman2022flow,
  title={Flow matching for generative modeling},
  author={Lipman, Yaron and Chen, Ricky TQ and Ben-Hamu, Heli and Nickel, Maximilian and Le, Matt},
  journal={arXiv preprint arXiv:2210.02747},
  year={2022}
}

@article{kaelbling1998planning,
  title={Planning and acting in partially observable stochastic domains},
  author={Kaelbling, Leslie Pack and Littman, Michael L and Cassandra, Anthony R},
  journal={Artificial intelligence},
  volume={101},
  number={1-2},
  pages={99--134},
  year={1998},
  publisher={Elsevier}
}

@article{lauri2022partially,
  title={Partially observable markov decision processes in robotics: A survey},
  author={Lauri, Mikko and Hsu, David and Pajarinen, Joni},
  journal={IEEE Transactions on Robotics},
  volume={39},
  number={1},
  pages={21--40},
  year={2022},
  publisher={IEEE}
}

@article{fu2024mobile,
  title={Mobile aloha: Learning bimanual mobile manipulation with low-cost whole-body teleoperation},
  author={Fu, Zipeng and Zhao, Tony Z and Finn, Chelsea},
  journal={arXiv preprint arXiv:2401.02117},
  year={2024}
}

@article{li2025simplevla,
  title={Simplevla-rl: Scaling vla training via reinforcement learning},
  author={Li, Haozhan and Zuo, Yuxin and Yu, Jiale and Zhang, Yuhao and Yang, Zhaohui and Zhang, Kaiyan and Zhu, Xuekai and Zhang, Yuchen and Chen, Tianxing and Cui, Ganqu and others},
  journal={arXiv preprint arXiv:2509.09674},
  year={2025}
}

@article{bjorck2025gr00t,
  title={Gr00t n1: An open foundation model for generalist humanoid robots},
  author={Bjorck, Johan and Casta{\~n}eda, Fernando and Cherniadev, Nikita and Da, Xingye and Ding, Runyu and Fan, Linxi and Fang, Yu and Fox, Dieter and Hu, Fengyuan and Huang, Spencer and others},
  journal={arXiv preprint arXiv:2503.14734},
  year={2025}
}

@article{intelligence2025pi_,
  title={pi0.5: a Vision-Language-Action Model with Open-World Generalization},
  author={Intelligence, Physical and Black, Kevin and Brown, Noah and Darpinian, James and Dhabalia, Karan and Driess, Danny and Esmail, Adnan and Equi, Michael and Finn, Chelsea and Fusai, Niccolo and others},
  journal={arXiv preprint arXiv:2504.16054},
  year={2025}
}

@article{zhang2023llama,
  title={Llama-adapter: Efficient fine-tuning of language models with zero-init attention},
  author={Zhang, Renrui and Han, Jiaming and Liu, Chris and Gao, Peng and Zhou, Aojun and Hu, Xiangfei and Yan, Shilin and Lu, Pan and Li, Hongsheng and Qiao, Yu},
  journal={arXiv preprint arXiv:2303.16199},
  year={2023}
}

@article{bai2025qwen2,
  title={Qwen2. 5-vl technical report},
  author={Bai, Shuai and Chen, Keqin and Liu, Xuejing and Wang, Jialin and Ge, Wenbin and Song, Sibo and Dang, Kai and Wang, Peng and Wang, Shijie and Tang, Jun and others},
  journal={arXiv preprint arXiv:2502.13923},
  year={2025}
}

@inproceedings{karamcheti2024prismatic,
  title={Prismatic vlms: Investigating the design space of visually-conditioned language models},
  author={Karamcheti, Siddharth and Nair, Suraj and Balakrishna, Ashwin and Liang, Percy and Kollar, Thomas and Sadigh, Dorsa},
  booktitle={Forty-first International Conference on Machine Learning},
  year={2024}
}

@article{beyer2024paligemma,
  title={Paligemma: A versatile 3b vlm for transfer},
  author={Beyer, Lucas and Steiner, Andreas and Pinto, Andr{\'e} Susano and Kolesnikov, Alexander and Wang, Xiao and Salz, Daniel and Neumann, Maxim and Alabdulmohsin, Ibrahim and Tschannen, Michael and Bugliarello, Emanuele and others},
  journal={arXiv preprint arXiv:2407.07726},
  year={2024}
}

@article{fan2024learning,
  title={Learning robust skills for tightly coordinated arms in contact-rich tasks},
  author={Fan, Yaowei and Li, Xinge and Zhang, Kaihang and Qian, Chen and Zhou, Fanghao and Li, Tiefeng and Huang, Zhilong},
  journal={IEEE Robotics and Automation Letters},
  volume={9},
  number={3},
  pages={2973--2980},
  year={2024},
  publisher={IEEE}
}

@article{cui2024task,
  title={A task-adaptive deep reinforcement learning framework for dual-arm robot manipulation},
  author={Cui, Yuanzhe and Xu, Zhipeng and Zhong, Lou and Xu, Pengjie and Shen, Yichao and Tang, Qirong},
  journal={IEEE Transactions on Automation Science and Engineering},
  volume={22},
  pages={466--479},
  year={2024},
  publisher={IEEE}
}

@article{huang20243d,
  title={3d-vitac: Learning fine-grained manipulation with visuo-tactile sensing},
  author={Huang, Binghao and Wang, Yixuan and Yang, Xinyi and Luo, Yiyue and Li, Yunzhu},
  journal={arXiv preprint arXiv:2410.24091},
  year={2024}
}

@article{lin2023bi,
  title={Bi-touch: Bimanual tactile manipulation with sim-to-real deep reinforcement learning},
  author={Lin, Yijiong and Church, Alex and Yang, Max and Li, Haoran and Lloyd, John and Zhang, Dandan and Lepora, Nathan F},
  journal={IEEE Robotics and Automation Letters},
  volume={8},
  number={9},
  pages={5472--5479},
  year={2023},
  publisher={IEEE}
}

@inproceedings{stepputtis2022system,
  title={A system for imitation learning of contact-rich bimanual manipulation policies},
  author={Stepputtis, Simon and Bandari, Maryam and Schaal, Stefan and Amor, Heni Ben},
  booktitle={2022 IEEE/RSJ International Conference on Intelligent Robots and Systems (IROS)},
  pages={11810--11817},
  year={2022},
  organization={IEEE}
}

@article{bahety2024screwmimic,
  title={Screwmimic: Bimanual imitation from human videos with screw space projection},
  author={Bahety, Arpit and Mandikal, Priyanka and Abbatematteo, Ben and Mart{\'\i}n-Mart{\'\i}n, Roberto},
  journal={arXiv preprint arXiv:2405.03666},
  year={2024}
}

@article{zhou2025you,
  title={You Only Teach Once: Learn One-Shot Bimanual Robotic Manipulation from Video Demonstrations},
  author={Zhou, Huayi and Wang, Ruixiang and Tai, Yunxin and Deng, Yueci and Liu, Guiliang and Jia, Kui},
  journal={arXiv preprint arXiv:2501.14208},
  year={2025}
}

@article{zhou2025binomap,
  title={BiNoMaP: Learning Category-Level Bimanual Non-Prehensile Manipulation Primitives},
  author={Zhou, Huayi and Jia, Kui},
  journal={arXiv preprint arXiv:2509.21256},
  year={2025}
}

@article{zhao2024aloha,
  title={Aloha unleashed: A simple recipe for robot dexterity},
  author={Zhao, Tony Z and Tompson, Jonathan and Driess, Danny and Florence, Pete and Ghasemipour, Kamyar and Finn, Chelsea and Wahid, Ayzaan},
  journal={arXiv preprint arXiv:2410.13126},
  year={2024}
}

@article{gkanatsios20253d,
  title={3D FlowMatch Actor: Unified 3D Policy for Single-and Dual-Arm Manipulation},
  author={Gkanatsios, Nikolaos and Xu, Jiahe and Bronars, Matthew and Mousavian, Arsalan and Ke, Tsung-Wei and Fragkiadaki, Katerina},
  journal={arXiv preprint arXiv:2508.11002},
  year={2025}
}

@article{Yang2024Qwen25TR,
  title={Qwen2.5 Technical Report},
  author={Qwen An Yang and Baosong Yang and Beichen Zhang and Binyuan Hui and Bo Zheng and Bowen Yu and Chengyuan Li and Dayiheng Liu and Fei Huang and Guanting Dong and Haoran Wei and Huan Lin and Jian Yang and Jianhong Tu and Jianwei Zhang and Jianxin Yang and Jiaxin Yang and Jingren Zhou and Junyang Lin and Kai Dang and Keming Lu and Keqin Bao and Kexin Yang and Le Yu and Mei Li and Mingfeng Xue and Pei Zhang and Qin Zhu and Rui Men and Runji Lin and Tianhao Li and Tingyu Xia and Xingzhang Ren and Xuancheng Ren and Yang Fan and Yang Su and Yi-Chao Zhang and Yunyang Wan and Yuqi Liu and Zeyu Cui and Zhenru Zhang and Zihan Qiu and Shanghaoran Quan and Zekun Wang},
  journal={ArXiv},
  year={2024},
  volume={abs/2412.15115},
}

@article{dinov2,
  author       = {Maxime Oquab and
                  Timoth{\'{e}}e Darcet and
                  Th{\'{e}}o Moutakanni and
                  Huy V. Vo and
                  Marc Szafraniec and
                  Vasil Khalidov and
                  Pierre Fernandez and
                  Daniel Haziza and
                  Francisco Massa and
                  Alaaeldin El{-}Nouby and
                  Mido Assran and
                  Nicolas Ballas and
                  Wojciech Galuba and
                  Russell Howes and
                  Po{-}Yao Huang and
                  Shang{-}Wen Li and
                  Ishan Misra and
                  Michael Rabbat and
                  Vasu Sharma and
                  Gabriel Synnaeve and
                  Hu Xu and
                  Herv{\'{e}} J{\'{e}}gou and
                  Julien Mairal and
                  Patrick Labatut and
                  Armand Joulin and
                  Piotr Bojanowski},
  title        = {DINOv2: Learning Robust Visual Features without Supervision},
  journal      = {Trans. Mach. Learn. Res.},
  volume       = {2024},
  year         = {2024}
}

@inproceedings{zhai2023sigmoid,
  title={Sigmoid loss for language image pre-training},
  author={Zhai, Xiaohua and Mustafa, Basil and Kolesnikov, Alexander and Beyer, Lucas},
  booktitle={Proceedings of the IEEE/CVF international conference on computer vision},
  pages={11975--11986},
  year={2023}
}

@article{huang2025otter,
  title={OTTER: A Vision-Language-Action Model with Text-Aware Visual Feature Extraction},
  author={Huang, Huang and Liu, Fangchen and Fu, Letian and Wu, Tingfan and Mukadam, Mustafa and Malik, Jitendra and Goldberg, Ken and Abbeel, Pieter},
  journal={arXiv preprint arXiv:2503.03734},
  year={2025}
}

@article{song2025reconvla,
  title={ReconVLA: Reconstructive Vision-Language-Action Model as Effective Robot Perceiver},
  author={Song, Wenxuan and Zhou, Ziyang and Zhao, Han and Chen, Jiaxin and Ding, Pengxiang and Yan, Hang and Huang, Yan and Tang, Fan and Wang, Donglin and Li, Hongsheng},
  journal={arXiv preprint arXiv:2508.10333},
  year={2025}
}

@article{li2026dtp,
  title={{DTP}: A Simple yet Effective Distracting Token Pruning Framework for Vision-Language Action Models},
  author={Li, Chenyang and Liu, Jieyuan and Li, Bin and Gao, Bo and Yuan, Yilin and He, Yangfan and Li, Yuchen and Tang, Jingqun},
  journal={arXiv preprint arXiv:2601.16065},
  year={2026},
  doi={10.48550/arXiv.2601.16065}
}

@article{li2026light,
  title={Light-WAM: Efficient World Action Models with State-Fusion Action Decoding},
  author={Li, Ziang and Cheng, Dongzhou and Wang, Yibin and Wang, Shiyue and Xu, Xiaoyang and Weng, Lingxuan and Wang, Juan and Wang, Jiaqi},
  journal={arXiv preprint arXiv:2606.08242},
  year={2026}
}

@article{srivastava2014dropout,
  title={Dropout: A simple way to prevent neural networks from overfitting},
  author={Srivastava, Nitish and Hinton, Geoffrey and Krizhevsky, Alex and Sutskever, Ilya and Salakhutdinov, Ruslan},
  journal={The Journal of Machine Learning Research},
  volume={15},
  number={1},
  pages={1929--1958},
  year={2014},
  publisher={JMLR.org}
}

@inproceedings{huang2016deep,
  title={Deep networks with stochastic depth},
  author={Huang, Gao and Sun, Yu and Liu, Zhuang and Sedra, Daniel and Weinberger, Kilian Q},
  booktitle={European Conference on Computer Vision},
  pages={646--661},
  year={2016},
  organization={Springer}
}

@inproceedings{vaswani2017attention,
  title={Attention is all you need},
  author={Vaswani, Ashish and Shazeer, Noam and Parmar, Niki and Uszkoreit, Jakob and Jones, Llion and Gomez, Aidan N and Kaiser, {\L}ukasz and Polosukhin, Illia},
  booktitle={Advances in Neural Information Processing Systems},
  volume={30},
  year={2017}
}

@article{neverova2016moddrop,
  title={ModDrop: Adaptive multi-modal gesture recognition},
  author={Neverova, Natalia and Wolf, Christian and Taylor, Graham and Nebout, Florian},
  journal={IEEE Transactions on Pattern Analysis and Machine Intelligence},
  volume={38},
  number={8},
  pages={1692--1706},
  year={2016},
  publisher={IEEE}
}

@inproceedings{liu2017sensor,
  title={Sensor dropout: Learning efficient sensing policies},
  author={Liu, Yan and Pardo, Fabio and Sch{\"o}n, Thomas B},
  booktitle={Workshop on Bayesian Deep Learning, NeurIPS},
  year={2017}
}

\clearpage
\appendix
\section{Experimental Settings}
\label{supp:setting}

\begin{table}[tb]
    \centering
    \caption{VRAM footprint during simulation on RTX 4090 (24\,GB). OpenVLA-OFT approaches the 24\,GB ceiling and frequently triggers OOM failures, whereas VLA-Adapter (0.5B) operates with a substantially smaller footprint, enabling stable iteration~\cite{wang2025vla}.}
    \label{tab:vram_consumption}
    \begingroup
    \footnotesize
    \setlength{\tabcolsep}{4pt}
    \begin{tabular}{@{}c|c|c@{}}
        \toprule
        Model & VRAM (GB) & Status \\
        \midrule
        OpenVLA-OFT (7B) & $\approx 23\text{-}24$ & OOM-prone \\
        VLA-Adapter (0.5B) & $\approx 10$ & Stable \\
        \bottomrule
    \end{tabular}
    \endgroup
\end{table}

\subsection{Backbone Rationale and Motivation}
We adopt VLA-Adapter~\cite{wang2025vla} as our primary backbone, motivated by its parameter efficiency and alignment with our research objectives.

\noindent\textbf{\textit{Parameter and Inference Efficiency.}} VLA-Adapter leverages a 0.5B-scale vision-language backbone with a lightweight bridging attention strategy, enabling parameter-efficient fine-tuning without large-scale robotic pretraining. This aligns well with our goal of enhancing bimanual manipulation under limited training resources. It also offers low-latency inference via one-shot decoding of learnable action queries.

\noindent\textbf{\textit{Practical and Computational Constraints.}} In contrast, the 7B-parameter OpenVLA-OFT~\cite{kim2025fine} demands more parameters and query embeddings for bimanual tasks. This imposes significant overhead in multi-view and long-horizon scenarios, causing high VRAM sensitivity. This burden is exacerbated during Clean2Rand evaluation, where diverse environment assets must be rendered concurrently. In our engineering setup, we simulate with an RTX 4090 GPU since the H100 lacks Vulkan support for SAPIEN rendering. Consequently, the combined VRAM footprint of OpenVLA-OFT and simulation rendering frequently approaches the 24GB ceiling (see Table~\ref{tab:vram_consumption}). This leads to frequent \textbf{Out-Of-Memory (OOM)} failures and impedes experimental throughput. Therefore, to support faster iteration on commodity-grade GPUs, we prioritize VLA-Adapter.

\noindent\textbf{\textit{Empirical Suitability in Simulation.}} This preference is not solely resource-driven. In the original VLA-Adapter paper, the authors report that VLA-Adapter achieves performance broadly comparable to OpenVLA-OFT across multiple simulated results, and in some cases slightly better, despite using a substantially smaller backbone~\cite{wang2025vla}. We observe a similar tendency in our own domain-clean evaluation: as summarized later in Table~\ref{tab:oft_performance_comparison}, the Adapter baseline remains competitive and attains a higher average success rate than OpenVLA-OFT. While these comparisons are drawn from different benchmarks and implementations, they suggest that selecting VLA-Adapter is a practically reasonable backbone choice rather than merely a compromise for memory efficiency.

\noindent\textbf{\textit{Methodological Motivation.}} Methodologically, we observe that query-based VLA models (e.g., VLA-Adapter) can exhibit unstable multi-view and language fusion in bimanual control, often coinciding with attention spreading to distracting regions. This observation motivates our proposed improvements. To this end, we introduce M3: a modality masking strategy effective exclusively during the training phase. Based on a heuristic design, M3 operates by stochastically masking partial modality channels and action query tokens. Without altering the baseline architecture, this approach provides structured partial observability during training and encourages more robust evidence use.

\subsection{Baselines.}
\paragraph{Action Chunking with Transformers (ACT)~\cite{zhao2023learning}.} ACT, based on a conditional variational autoencoder (CVAE) imitation learning formulation, uses action chunking to predict future action sequences and applies temporal ensembling techniques to ensure stable and smooth execution. Inspired by DETR, the model adopts an encoder–decoder architecture where action queries feed into the transformer decoder to predict actions. It trains through direct behavior cloning with a regression loss applied to complete action chunks, and empirically demonstrates the ability to solve fine-grained bimanual manipulation tasks on low-cost hardware.

\paragraph{Diffusion Policy (DP)~\cite{chi2025diffusion}.} Diffusion Policy treats visuomotor control as a conditional denoising diffusion process over action trajectories. It learns to iteratively refine noisy action sequences into expert‑like control signals conditioned on current observations. The forward process adds Gaussian noise to ground‑truth action horizons, while a neural network learns the reverse denoising dynamics given images and robot state. At inference, the model samples a full future action horizon via iterative denoising and executes only the initial segment before re‑planning. This enables robust receding‑horizon control and naturally captures multimodal action distributions. Empirically, it outperforms standard behavioral cloning across many manipulation tasks and serves as the canonical “diffusion over actions” formulation that many VLA architectures adopt as their continuous‑action head or expert.

\paragraph{Robotics Diffusion Transformer (RDT)~\cite{liu2024rdt}.} Addressing the high complexity and data scarcity of bimanual manipulation, RDT-1B presents itself as a 1.2B parameter diffusion-based foundation model. The work introduces a "Physically Interpretable Unified Action Space" to unify action formats, which facilitates the pre-training of its Robotics Diffusion Transformer (RDT) on a massive 1M+ multi-robot trajectory dataset. The model subsequently undergoes fine-tuning on 6K+ bimanual-specific episodes. By leveraging a scalable Transformer and diffusion modeling to represent multi-modal action distributions, RDT-1B achieves strong zero-shot generalization to new objects and scenes, follows language instructions, and learns new skills from just 1-5 demonstrations.

\paragraph{$\pi_{0}$~\cite{black2024pi_0}.} This model represents a large Vision-Language-Action (VLA) foundation policy built on a pretrained Vision-Language Model (VLM)~\cite{beyer2024paligemma}, augmented with a continuous-action flow-matching~\cite{lipman2022flow} expert to enable generalist robot control. The VLM encodes images and natural-language instructions, inheriting internet-scale semantic knowledge, while the flow model learns a conditional vector field that transforms noise into continuous robot actions given visual, linguistic, and proprioceptive context. $\pi_0$ trains on a variety of datasets for single-arm, bimanual, and mobile manipulators, follows complex language instructions, and allows for fine-tuning to acquire new skills. As a VLM-based prototype VLA foundation model, it influences a large body of subsequent diffusion-based VLA work.

\paragraph{VLA-Adapter (Adapter)~\cite{wang2025vla}.} VLA-Adapter builds a VLA by freezing a 0.5B vision-language backbone and learning lightweight adapter and policy modules that bridge multimodal features to actions. Rather than pre-training the VLM on robot data, it identifies the most control-relevant vision-language conditions and injects them into the action space through a policy module with Bridge Attention. With modest amounts of robotic data, VLA-Adapter reaches competitive performance on simulated and real benchmarks while retaining fast inference and much lower training cost than billion-parameter VLAs. It is therefore a representative small-scale VLA baseline centered on efficient VL-to-action bridging instead of backbone scaling or heavy diffusion or flow heads.

\subsection{Additional Configuration}
\label{sec:more_details}

All models are trained on a high-performance compute cluster equipped with 4 $\times$ NVIDIA H100 GPUs to ensure optimization efficiency, while all evaluations and inference benchmarks are conducted on a single NVIDIA RTX 4090 GPU. We maintain consistent training hyperparameters between the VLA-Adapter baseline and our M3 method, employing a global batch size of 48 (12 per GPU). For the OpenVLA-OFT experiments, we adjust the global batch size to 16 (4 per GPU) to accommodate the higher memory footprint of the 7B backbone, while keeping all other training configurations identical to those of the VLA-Adapter. Furthermore, adhering to the standard dual-arm Aloha protocols~\cite{zhao2023learning} established in RoboTwin 2.0 and OpenVLA-OFT, we strictly adopt an action chunk size of 25 steps across all training and evaluation sessions.

\subsection{Tasks.}
\begin{table}[tb!]
  \centering
  \caption{Task statistics and horizon categorization on the RoboTwin 2.0 benchmark.}
\resizebox{0.5\linewidth}{!}{
  \begin{tabular}{c|c|c|c}
    \toprule
    \textbf{Task Name} & \textbf{Steps} & \textbf{Horizon} & \textbf{Horizon Group} \\
    \midrule
    \multicolumn{4}{c}{\cellcolor{green!20}\textbf{Short Horizon Tasks}} \\
    \midrule
    Click Bell & 62 & Short & \multirow{3}{*}{\makecell{Average: 95 steps \\Count: 3 tasks}} \\
    Grab Roller & 95 & Short & \\
    Place Phone Stand & 128 & Short & \\
    \midrule
    \multicolumn{4}{c}{\cellcolor{yellow!20}\textbf{Medium Horizon Tasks}} \\
    \midrule
    Place Bread Basket & 222 & Medium & \multirow{4}{*}{\makecell{Average: 197 steps \\Count: 4 tasks}} \\
    Place A2B Right & 140 & Medium & \\
    Place Shoe & 149 & Medium & \\
    Stack Blocks Two & 277 & Medium & \\
    \midrule
    \multicolumn{4}{c}{\cellcolor{orange!20}\textbf{Long Horizon Tasks}} \\
    \midrule
    Handover Block & 287 & Long & \multirow{3}{*}{\makecell{Average: 461 steps \\Count: 3 tasks}} \\
    Put Bottles Dustbin & 592 & Long & \\
    Block Rank Size & 505 & Long & \\
    \midrule
    \multicolumn{1}{l|}{\textbf{Overall Statistics}} & \multicolumn{3}{c}{\textbf{Total: 10 tasks, Average: 245 steps.}} \\
    \bottomrule
  \end{tabular}
  }
  \label{tab:task_horizon_analysis}
\end{table}

\begin{table*}[t]
\centering
\small 
\vspace{0.3cm}
\caption{Representative seen (training) and unseen (evaluation) language instructions. During evaluation, instructions are stochastically sampled from the complete unseen set.}
\setlength{\tabcolsep}{6pt} 
\renewcommand{\arraystretch}{1.15}

\begin{tabularx}{\textwidth}{c c  X X} 
\toprule
\textbf{Category} & \textbf{Task Name} & \textbf{Seen Instructions} & \textbf{Unseen Instructions} \\
\midrule

\multirow{6}{*}{Short Horizon} 
 & Click Bell & Touch the bell at its top center. & Find the palm-sized bell and click its top center. \\
 \cmidrule{3-4}
 & Grab Roller & Grab the roller on the table. & Use arms to grab the light brown cylindrical roller.\\
 \cmidrule{3-4}
 & Place Phone Stand & Move the palm-sized phone onto the compact plastic phonestand. & Set the handheld phone with silver frame on the adjustable phone holder.\\
\midrule

\multirow{10}{*}{Medium Horizon}
 & Place Bread Basket & Grab both the small bread and the golden bread loaf, drop into the beige plastic breadbasket. & Use one arm to grab the palm-sized braided golden bread, drop in the round beige breadbasket.\\
 \cmidrule{3-4}
 & Place A2B Right & Place the manual blue stapler precisely to the right of the woodenblock with back grooves. & Put the hard black box for card deck to the right of the bright blue soap bar.\\
 \cmidrule{3-4}
 & Place Shoe & Move the white shoe with rounded toe from the table to the mat in one fluid motion. & Grab the synthetic shoe upper from the table and set it on the mat. \\
 \cmidrule{3-4}
 & Stack Blocks Two & Shift red block to the center and place green block on top. & Place red block in the middle, then stack green block on it. \\
\midrule

\multirow{9}{*}{Long Horizon} 
 & Handover Block & Hold the red block with the left arm and place it on the pad. & Grab the red block with the left arm, switch to the right, and place it on the blue pad. \\
 \cmidrule{3-4}
 & Put Bottles Dustbin & Take the bottle made of smooth plastic, drop it in the dustbin with lid and bag, then move the blue-accented cylindrical bottle and the black and red bottle to the dustbin with lid and bag. & Grab the yellow bottle with elongated cylinder body and place it in the curved rectangular trash can, then do the same for the plastic bottle and the bottle with red screw cap.\\
 \cmidrule{3-4}
 & Block Rank Size & Bring large block, medium block, and small block to the center, sorting largest to smallest. & Sort the three blocks by size at the center, from the largest block to the smallest. \\

\bottomrule
\end{tabularx}

\label{tab:instructions}
\end{table*}

\begin{figure*}[t!]
  \centering
  \includegraphics[width=.95\textwidth]{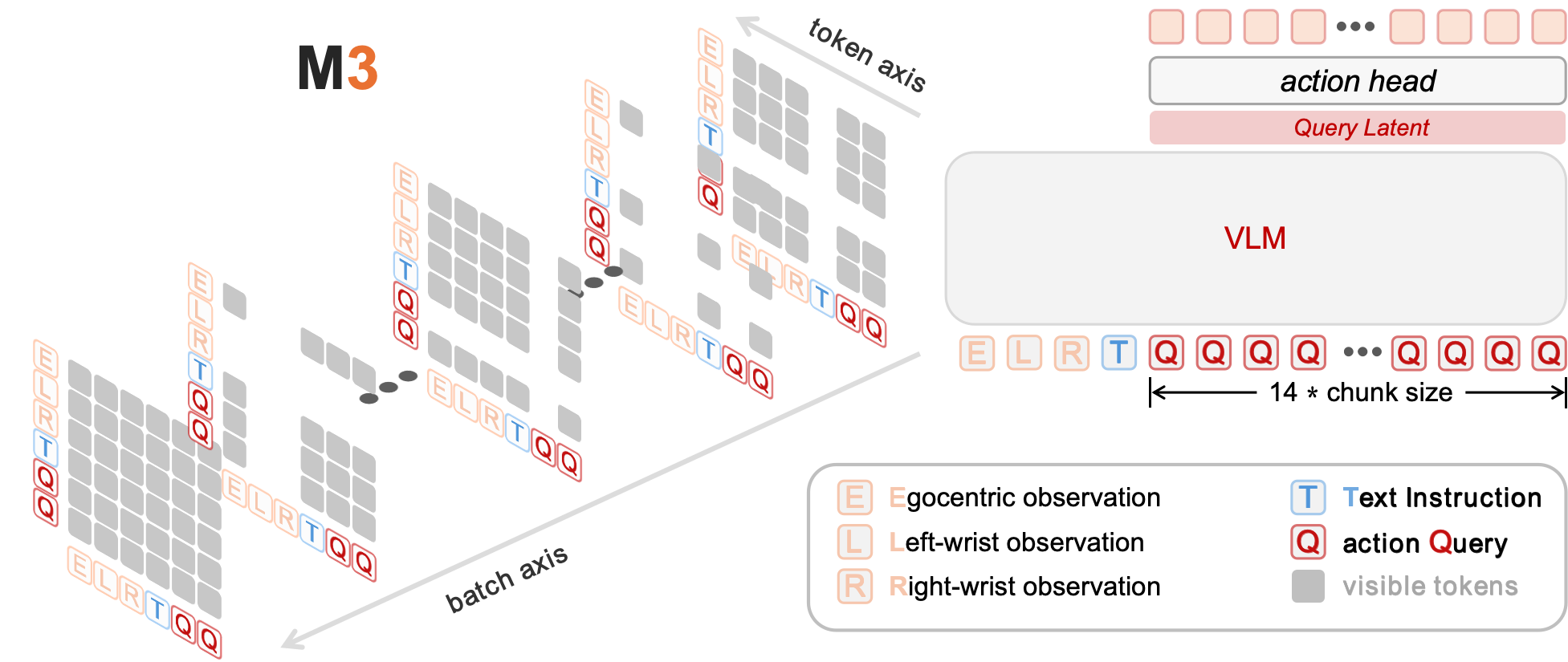}
\caption{Overview of the Modality Masking Mechanism (M3) for OpenVLA-OFT~\cite{kim2025fine}, which is built upon the OpenVLA-7B~\cite{kim2024openvla} backbone. The legend follows the same color and token conventions as the main-paper overview figure. The three main differences from VLA-Adapter~\cite{wang2025vla} include: (1) the adoption of parallel decoding with bidirectional attention, (2) the input of zero-query embeddings with shape $(\text{action axes}) \times (\text{chunk size})$, and (3) unlike VLA-Adapter, which interacts vision and query latents via an action expert at every layer, here the final-layer query latents are mapped pointwise to the minimal unit of the action signal. For example, if we infer $T=25$ action steps and each bimanual timestep has $A=14$ action axes, we provide $A \times T = 14 \times 25 = 350$ query embeddings.}
  \Description{A schematic adapts M3 to OpenVLA-OFT. It shows multimodal tokens and zero-query embeddings entering the OpenVLA-OFT backbone with masking applied during training, followed by pointwise mapping from final-layer query latents to action outputs.}

  \label{fig:oft_overview}
  \vspace{10pt}
\end{figure*}

To evaluate our proposed M3 strategy, we conduct experiments on the RoboTwin 2.0 benchmark~\cite{chen2025robotwin}. Following standard protocols, we select a set of 10 bimanual manipulation tasks that involve varying degrees of coordination, contact-rich interaction, and spatiotemporal reasoning.

\noindent\textbf{Task Horizons and Complexity.} 
As detailed in Table~\ref{tab:task_horizon_analysis}, we categorize these tasks into three groups based on their planning horizon and average step count to analyze performance across different temporal complexities:

\begin{itemize}
    \item \textbf{Short Horizon:} This category includes tasks such as \textit{Click Bell}, \textit{Grab Roller}, and \textit{Place Phone Stand}. These tasks typically involve fundamental reaching or grasping primitives with an average duration of approximately 95 steps.
    
    \item \textbf{Medium Horizon:} Tasks such as \textit{Place Bread Basket}, \textit{Place A2B Right}, \textit{Place Shoe}, and \textit{Stack Blocks Two} fall into this category. They require sequential actions—such as pick-and-place operations or dual-arm coordination—averaging 197 steps in length.
    
    \item \textbf{Long Horizon:} The category comprises \textit{Handover Block}, \textit{Put Bottles Dustbin}, and \textit{Block Rank Size}. These tasks generally demand extended reasoning and multi-stage execution. For instance, \textit{Block Rank Size} involves sorting multiple objects, extending the horizon beyond 500 steps (avg. 461 steps).
\end{itemize}

This multi-horizon stratification allows us to examine M3 across varying temporal complexities, including longer sequences where we observe that the baseline query-based VLA can exhibit drift or instability.

\noindent\textbf{Instructions.} 
To assess the model's language grounding capability and robustness to linguistic perturbations, we follow the official RoboTwin 2.0 benchmark settings to evaluate performance under two instruction settings, as enumerated in Table~\ref{tab:instructions}:

\begin{enumerate}
    \item \textbf{Seen Instructions:} The set of instructions that are explicitly visible to the model during the training phase (e.g., ``\textit{Touch the bell at its top center}'').
    \item \textbf{Unseen Instructions:} Instructions that share the same semantic intent as the training commands but employ more varied and complex phrasing. These are used exclusively during evaluation to assess the model's robustness to linguistic diversity (e.g., ``\textit{Find the palm-sized bell...}'').
\end{enumerate}


\section{Additional Experiments}
\label{supp:add_exp}


To further examine transfer beyond the primary backbone studied in the main paper, we extend M3 to the OpenVLA-OFT~\cite{kim2025fine} framework, which features a distinct query-based architecture with parallel decoding and bidirectional attention, as illustrated in Fig.~\ref{fig:oft_overview}. As detailed in Table~\ref{tab:oft_performance_comparison}, this integration improves the average success rate by 21.3\% over the standard OpenVLA-OFT baseline in the domain-clean setting. The consistency of these gains across short-, medium-, and long-horizon tasks provides additional supporting evidence that M3 may transfer beyond VLA-Adapter in the evaluated setting.

\begin{table*}[t!]
  \caption{Domain-clean performance across multi-horizon tasks on the \textbf{RoboTwin~2.0} simulation platform. 
  The columns labeled \textbf{+M3} denote the integration of our M3 training strategy into the \textit{preceding} baseline (Adapter and OpenVLA-OFT, respectively). The best performance in each row is bolded, and the second-best is underlined. $\Delta^{\text{oft}}$ denotes the relative improvement of \textbf{M3} over the OpenVLA-OFT baseline. Overall, these results provide additional supporting evidence that M3 may transfer beyond the primary VLA-Adapter backbone in the evaluated domain-clean setting.}
  \label{tab:oft_performance_comparison}
  \centering
  \setlength{\tabcolsep}{12pt}
  \resizebox{\textwidth}{!}{%
  \begin{tabular}{@{}c|c|cc|cc|cc|c@{}}
    \toprule
    Category & Task Name & RDT$^{*}$ & $\pi_0^{*}$ & Adapter & \cellcolor{gray!15}\textbf{+M3} & OpenVLA-OFT & \cellcolor{gray!15}\textbf{+M3} & $\Delta^{\text{oft}}$ \\
    \midrule
    \multirow{3}{*}{Short Horizon} 
      & Click Bell & 80 & 44 & 84 & \cellcolor{gray!15}\underline{97} & 86 & \cellcolor{gray!15}\textbf{100} & \textcolor{red!70!black}{+14} \\ 
      & Grab Roller & 74 & \underline{96} & 88 & \cellcolor{gray!15}\underline{96} & 94 & \cellcolor{gray!15}\textbf{97} & \textcolor{red!70!black}{+3} \\ 
      & Place Phone Stand & 15 & 35 & 10 & \cellcolor{gray!15}\underline{55} & 24 & \cellcolor{gray!15}\textbf{56} & \textcolor{red!70!black}{+32} \\ 
    \midrule
    \multirow{4}{*}{Medium Horizon}
      & Place Bread Basket & 10 & \underline{17} & 11 & \cellcolor{gray!15}\textbf{22} & 3 & \cellcolor{gray!15}13 & \textcolor{red!70!black}{+10} \\ 
      & Place A2B Right & 1 & \underline{27} & 4 & \cellcolor{gray!15}\textbf{28} & 8 & \cellcolor{gray!15}10 & \textcolor{red!70!black}{+2} \\ 
      & Place Shoe & 35 & 28 & 34 & \cellcolor{gray!15}\textbf{63} & 17 & \cellcolor{gray!15}\underline{50} & \textcolor{red!70!black}{+33} \\
      & Stack Blocks Two & 21 & 42 & \underline{78} & \cellcolor{gray!15}\textbf{83} & 22 & \cellcolor{gray!15}69 & \textcolor{red!70!black}{+47} \\ 
    \midrule
    \multirow{3}{*}{Long Horizon} 
      & Handover Block & \underline{45} & \underline{45} & 27 & \cellcolor{gray!15}\textbf{74} & 28 & \cellcolor{gray!15}39 & \textcolor{red!70!black}{+11} \\ 
      & Put Bottles Dustbin & 21 & 54 & 60 & \cellcolor{gray!15}\textbf{81} & 40 & \cellcolor{gray!15}\underline{75} & \textcolor{red!70!black}{+35} \\ 
      & Block Rank Size & 0 & 7 & 14 & \cellcolor{gray!15}\textbf{28} & 0 & \cellcolor{gray!15}\underline{26} & \textcolor{red!70!black}{+26} \\ 
    \midrule
  \multicolumn{2}{c|}{\textbf{Overall Avg}} 
    & 30.2 & 39.5 & 41.0 & \cellcolor{gray!15}\textbf{62.7} & 32.2 & \cellcolor{gray!15}\underline{53.5} & \textcolor{red!70!black}{+21.3} \\
    \bottomrule
  \end{tabular}
  }
\end{table*}

\subsection{Quantitative Attention-Misalignment Analysis}
\label{supp:attn_metric}

To complement the qualitative heatmaps shown in the main paper, we further analyze whether action discontinuities are statistically associated with a larger amount of attention allocated away from prompt-relevant visual patches. Following the analysis style of recent empirical studies on visuomotor attention quality~\cite{li2026dtp,huang2025otter,song2025reconvla}, we revisit the three representative tasks used in our ablation study, namely \textit{Place Phone Stand}, \textit{Place Shoe}, and \textit{Handover Block}. For each task, we recollect both temporally smooth rollouts and discontinuous rollouts from the evaluation logs and compute a DTP-style \emph{unimportant attention} statistic from the model's internal self-attention tensors~\cite{li2026dtp}.

Concretely, we first form an important patch set $G$ on the concatenated three-view visual tokens using prompt-to-visual relevance, and then sum the action-to-visual attention mass outside $G$. In the configuration used here, prompt-to-visual relevance is averaged over the last 12 transformer layers and $G$ retains the top 50\% visual patches by relevance. This post-hoc statistic does not affect action prediction.

\begin{figure}[t!]
  \centering  
  \includegraphics[width=.44\textwidth]{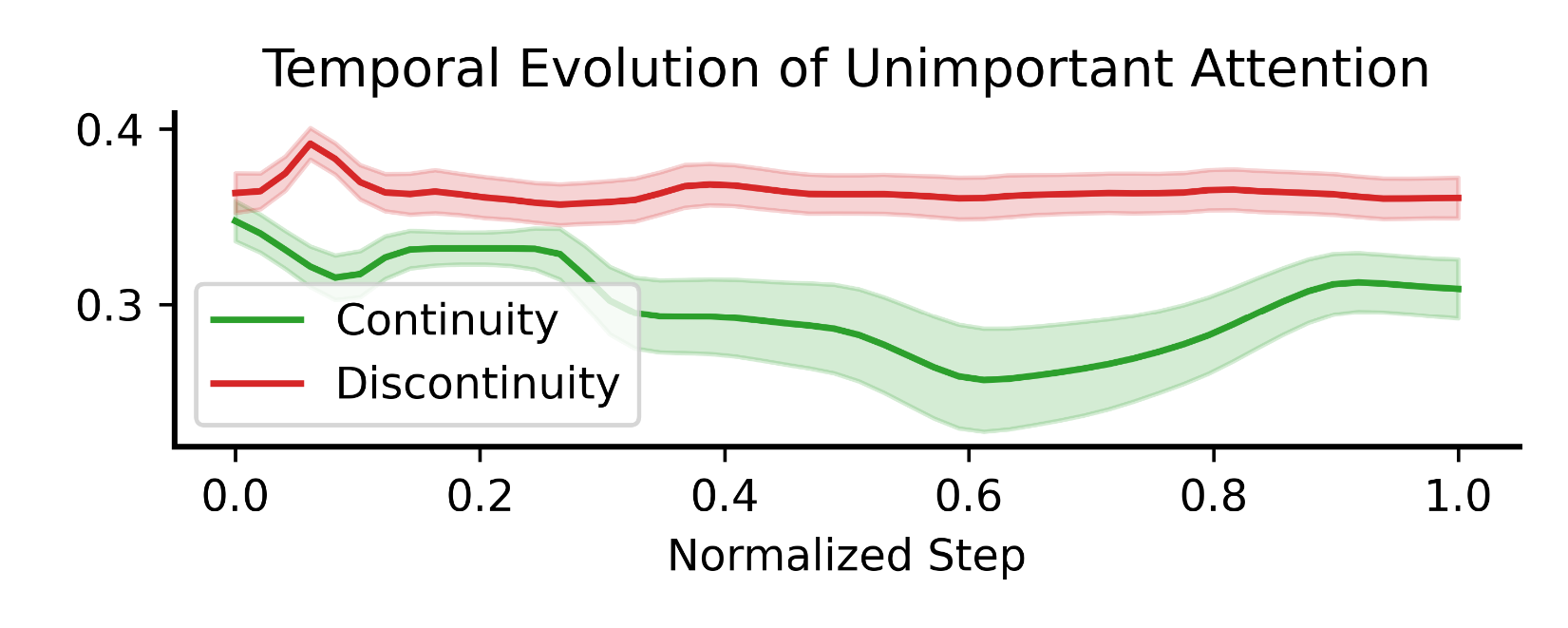}
  \captionsetup{hypcap=false}
  \captionof{figure}{Quantitative attention-misalignment analysis on three representative tasks. Discontinuous trajectories tend to exhibit higher unimportant attention than smooth trajectories over normalized rollout time.}
  \label{fig:misalignment_evidence}
\end{figure}

Because these trajectories have different horizons, we normalize each rollout to the interval $[0,1]$ and aggregate the per-step attention statistic along normalized time. The resulting comparison is shown in Fig.~\ref{fig:misalignment_evidence}. Across all three tasks, discontinuous trajectories tend to exhibit higher unimportant-attention values than their smooth counterparts. This gap is more visible in the middle-to-late phases of execution, where contact formation, object transfer, and final placement require accurate localization and temporally stable evidence selection. The trajectories produced by M3 tend to maintain a lower level of attention outside the prompt-relevant patch set throughout the rollout, suggesting that the policy may be less prone to drift toward distractors once the manipulation enters contact-rich stages.

This result provides a quantitative complement to the visualizations in the main paper: rollouts with larger action attention outside the prompt-relevant patch set tend to be more jittery or discontinuous. Together with the main-paper heatmaps, this trend is consistent with the hypothesis that training-time masking may reduce reliance on salient but task-irrelevant cues during cross-view fusion.

\noindent\textbf{Bimanual-specific interpretation.}
The above trend is particularly relevant to dual-arm manipulation. In single-arm settings, the wrist camera typically remains aligned with the acting hand and the egocentric view, so visual evidence from different viewpoints is less likely to compete. In our bimanual setting, however, the idle arm can introduce a distractor wrist view that remains visually salient but action-irrelevant for the current sub-step. This creates a characteristic form of cross-view interference that is weaker or absent in single-arm manipulation. Our masking rules, especially preserving the egocentric stream while jointly masking the wrist views, were designed to address this hypothesized failure mode and may help explain the generalization gains observed at inference. The trend in Fig.~\ref{fig:misalignment_evidence} is consistent with this interpretation.

\section{Visualization Results}
\label{supp:vis}

\begin{table}[!t]
  \centering
  \caption{Per-phase and full-task success rates (\%) for three real-world bimanual tasks, pooled over three evaluation rounds (clean: $3\!\times\!16\!=\!48$ trials; OOD: $3\!\times\!8\!=\!24$ trials per task). Full-task rows report the pooled rate together with the sample standard deviation across the three round-wise full-task success rates.}
  \label{tab:realworld_phase_multiround}
  \setlength{\tabcolsep}{4pt}
  \renewcommand{\arraystretch}{1.15}
  \small
  \begin{tabular}{@{} c l  cc  cc @{}}
    \toprule
    & & \multicolumn{2}{c}{\textbf{Clean}} & \multicolumn{2}{c}{\textbf{OOD}} \\
    \cmidrule(lr){3-4} \cmidrule(lr){5-6}
    \textbf{Task} & \textbf{Phase} & Adapter & \textbf{M3} & Adapter & \textbf{M3} \\
    \midrule
    \multirow{5}{*}{\rotatebox[origin=c]{90}{\scriptsize Bottle Cleanup}}
      & Brown-bottle disposal & 68.8 & 85.4 & 20.8 & 70.8 \\
      & Green-bottle lift     & 85.4 & 91.7 & 37.5 & 75.0 \\
      & Bimanual handover     & 54.2 & 79.2 & 16.7 & 62.5 \\
      & Final disposal        & 54.2 & 79.2 & 16.7 & 62.5 \\
      \cmidrule(l){2-6}
      & \textbf{Full task}    & 41.7{\scriptsize\,$\pm$9.6}  & \textbf{66.7}{\scriptsize\,$\pm$3.6}  & 16.7{\scriptsize\,$\pm$7.2}  & \textbf{58.3}{\scriptsize\,$\pm$19.1} \\
    \midrule
    \multirow{5}{*}{\rotatebox[origin=c]{90}{\scriptsize Stack \& Shelf}}
      & Bowl pickup           & 87.5 & 87.5 & 20.8 & 79.2 \\
      & Bowl stacking         & 45.8 & 75.0 & 16.7 & 70.8 \\
      & Stacked-bowl pickup   & 39.6 & 58.3 & 12.5 & 54.2 \\
      & Shelf placement       & 39.6 & 58.3 & 12.5 & 54.2 \\
      \cmidrule(l){2-6}
      & \textbf{Full task}    & 27.1{\scriptsize\,$\pm$9.6}  & \textbf{58.3}{\scriptsize\,$\pm$3.6}  & 12.5{\scriptsize\,$\pm$12.5} & \textbf{54.2}{\scriptsize\,$\pm$7.2} \\
    \midrule
    \multirow{5}{*}{\rotatebox[origin=c]{90}{\scriptsize Veggie Centering}}
      & Cucumber placement    & 91.7 & 97.9 & 25.0 & 87.5 \\
      & Eggplant placement    & 87.5 & 95.8 & 16.7 & 83.3 \\
      & Plate-rim grasp       & 75.0 & 89.6 & 12.5 & 70.8 \\
      & Plate centering       & 64.6 & 83.3 &  8.3 & 70.8 \\
      \cmidrule(l){2-6}
      & \textbf{Full task}    & 64.6{\scriptsize\,$\pm$3.6}  & \textbf{83.3}{\scriptsize\,$\pm$9.6}  &  8.3{\scriptsize\,$\pm$14.4} & \textbf{70.8}{\scriptsize\,$\pm$7.2} \\
    \bottomrule
  \end{tabular}
\end{table}

\noindent\textbf{Real-World stage-level trends and qualitative cases.}
Table~\ref{tab:realworld_phase_multiround} provides a stage-level breakdown of the same three-task real-world evaluation summarized in the main paper. The performance differences are distributed across multiple phases rather than concentrated in a single substep. In Bottle Cleanup, the largest clean-setting gaps appear at the bimanual handover and final disposal stages, where coordination demands are highest. In Stack \& Shelf, the two methods achieve comparable rates at the initial bowl pickup, while the gap is most pronounced at the bowl-stacking phase. In Veggie Centering, the gaps grow progressively from the early placement phases to the later plate-transport and centering stages. Under OOD clutter, baseline rates decline across all phases in each task, while M3 degrades comparatively less in this evaluation.

\begin{figure*}[t!]
  \centering
  \includegraphics[
    width=\textwidth,
    height=0.92\textheight,
    keepaspectratio
  ]{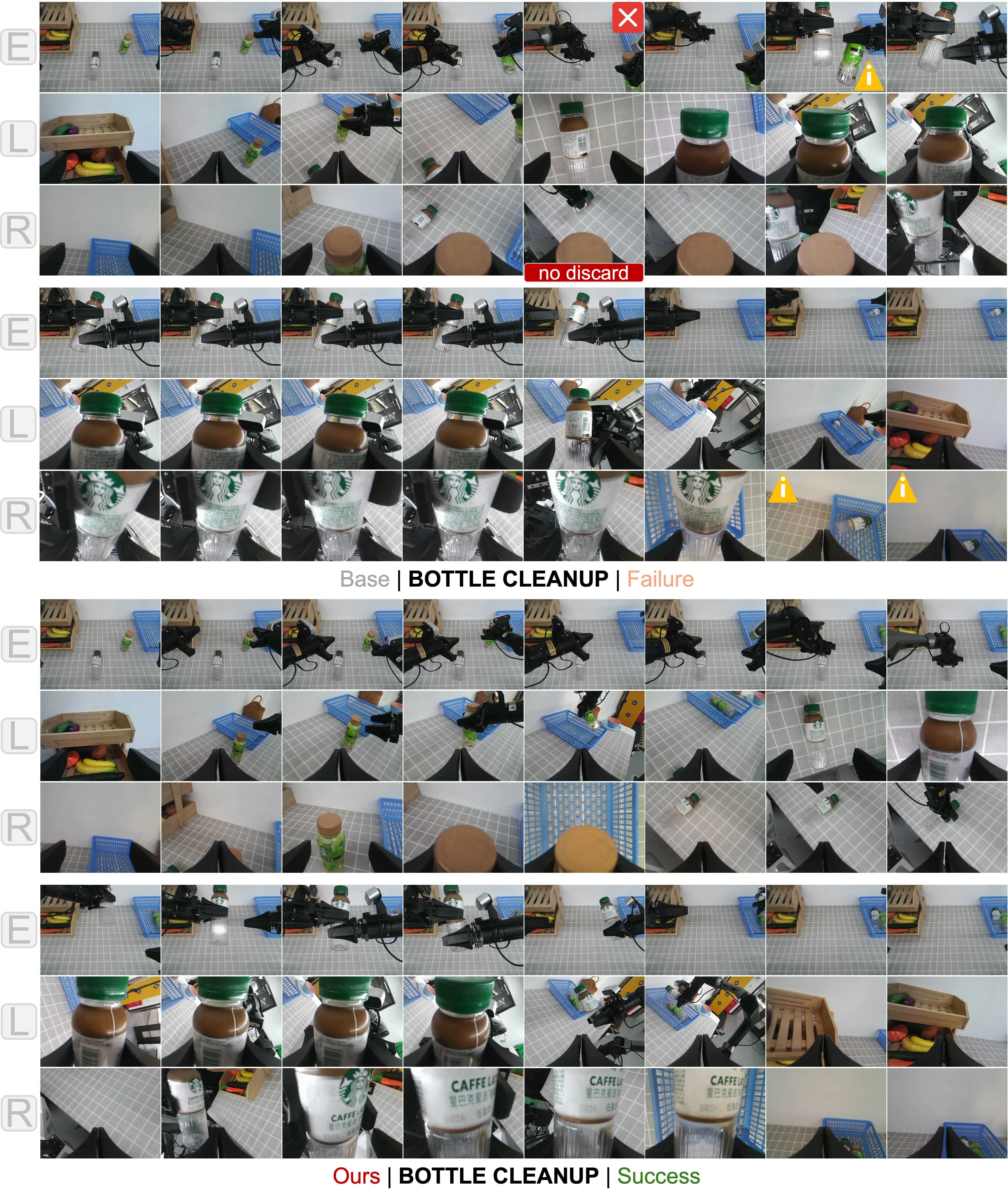}
  \caption{Bottle Cleanup: Case 1.}
  \Description{Real-world Bottle Cleanup rollout snapshots for Case 1.}
  \label{fig:sup_realworld_bottles_1}
\end{figure*}

\begin{figure*}[p]
  \centering
  \includegraphics[
    width=\textwidth,
    height=0.92\textheight,
    keepaspectratio
  ]{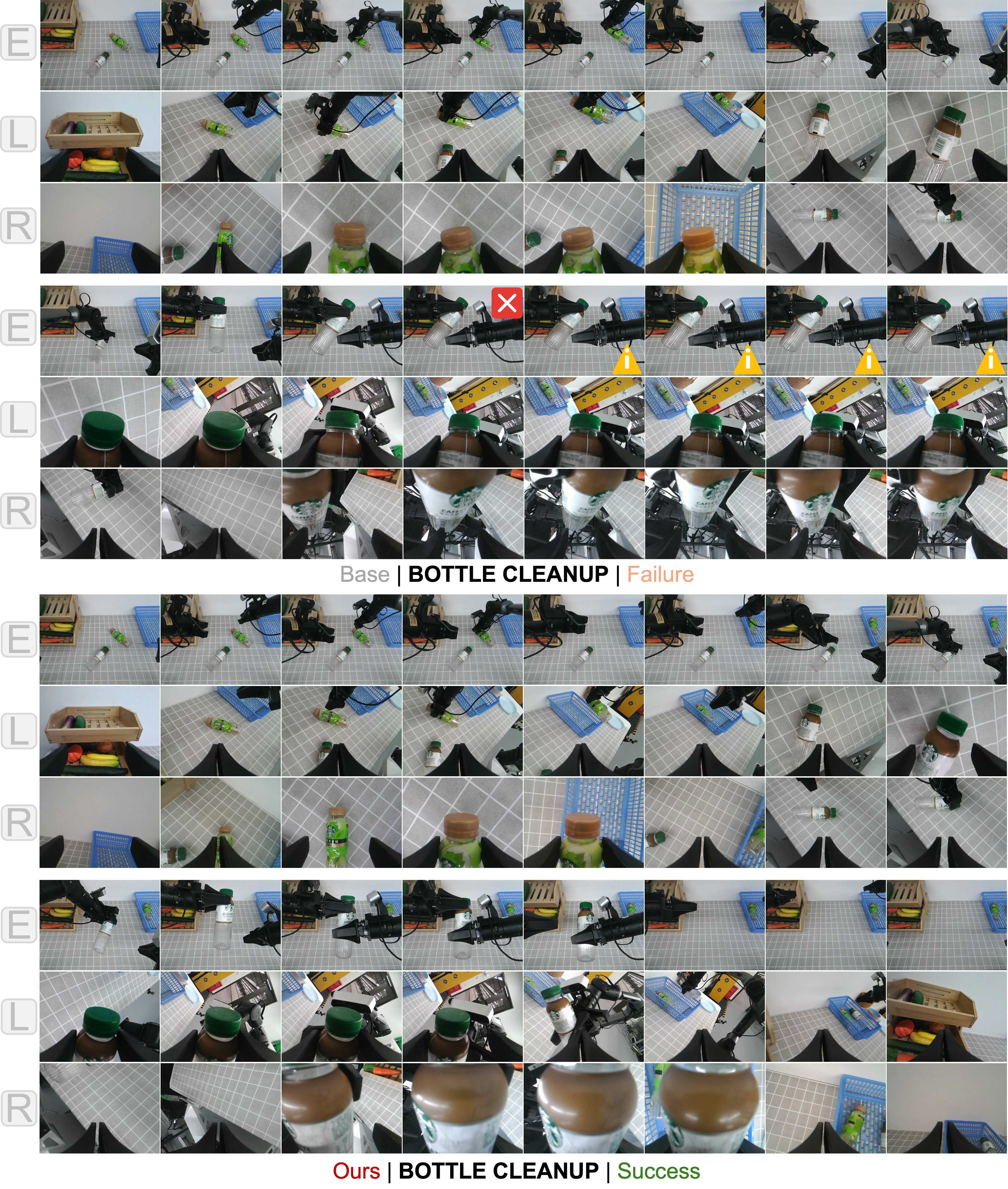}
  \caption{Bottle Cleanup: Case 2.}
  \Description{Real-world Bottle Cleanup rollout snapshots for Case 2.}
  \label{fig:sup_realworld_bottles_2}
\end{figure*}

\begin{figure*}[p]
  \centering
  \includegraphics[
    width=\textwidth,
    height=0.92\textheight,
    keepaspectratio
  ]{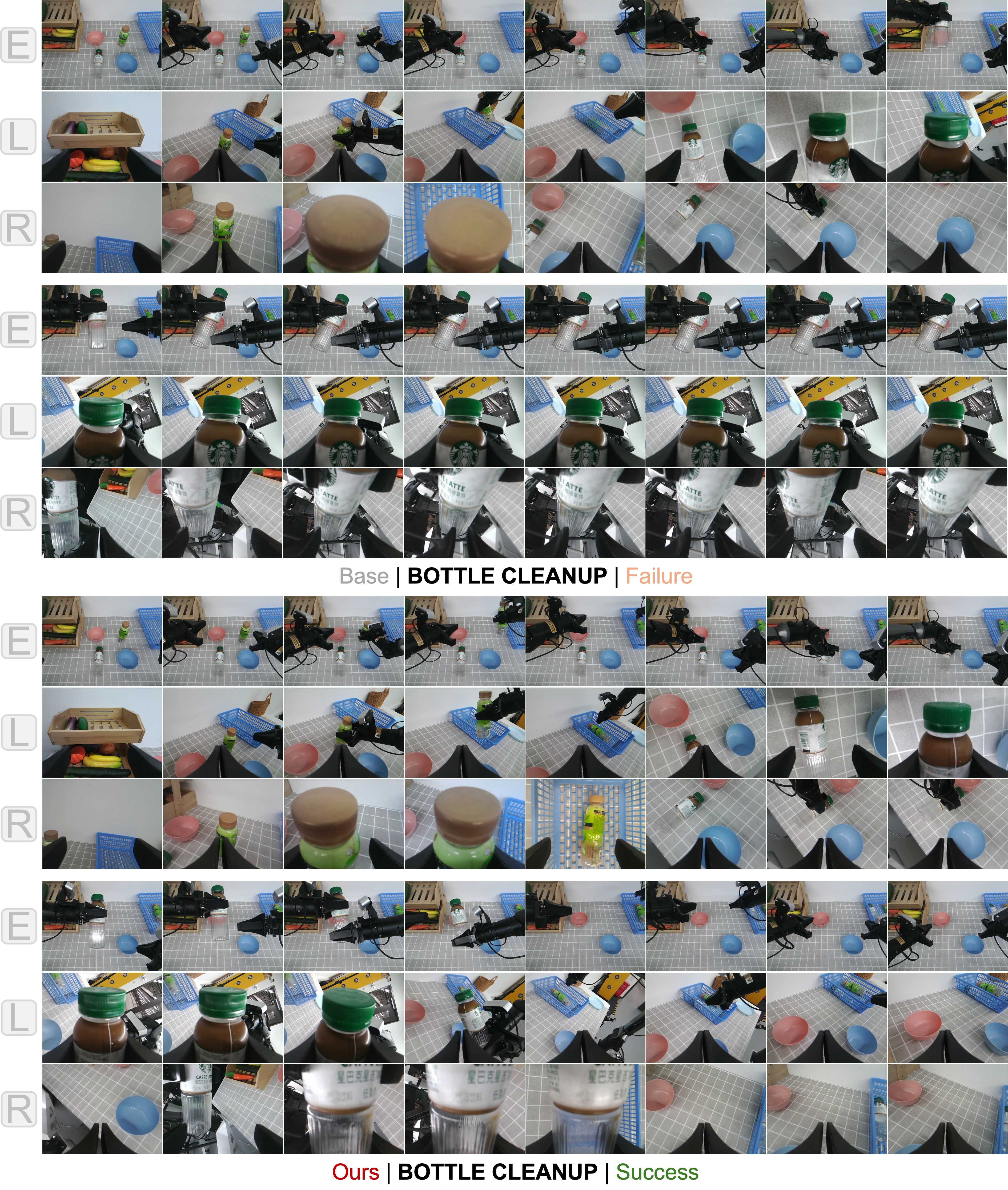}
  \caption{Bottle Cleanup: OOD Clutter.}
  \Description{Real-world Bottle Cleanup rollout snapshots for the OOD clutter case.}
  \label{fig:sup_realworld_bottles_ood}
\end{figure*}

\begin{figure*}[p]
  \centering
  \includegraphics[
    width=\textwidth,
    height=0.92\textheight,
    keepaspectratio
  ]{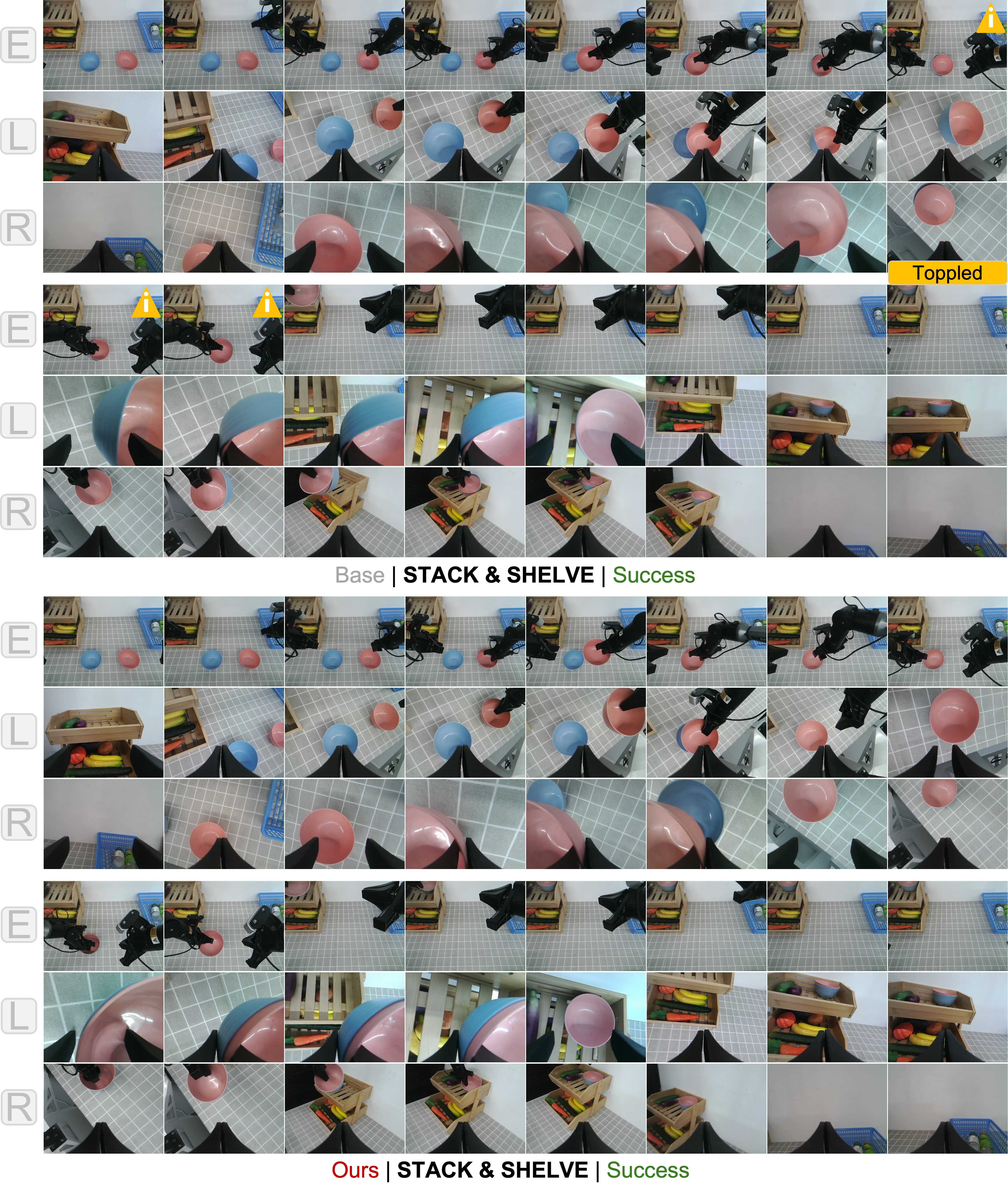}
  \caption{Stack \& Shelf: Case 1.}
  \Description{Real-world Stack and Shelf rollout snapshots for Case 1.}
  \label{fig:sup_realworld_stack_1}
\end{figure*}

\begin{figure*}[p]
  \centering
  \includegraphics[
    width=\textwidth,
    height=0.92\textheight,
    keepaspectratio
  ]{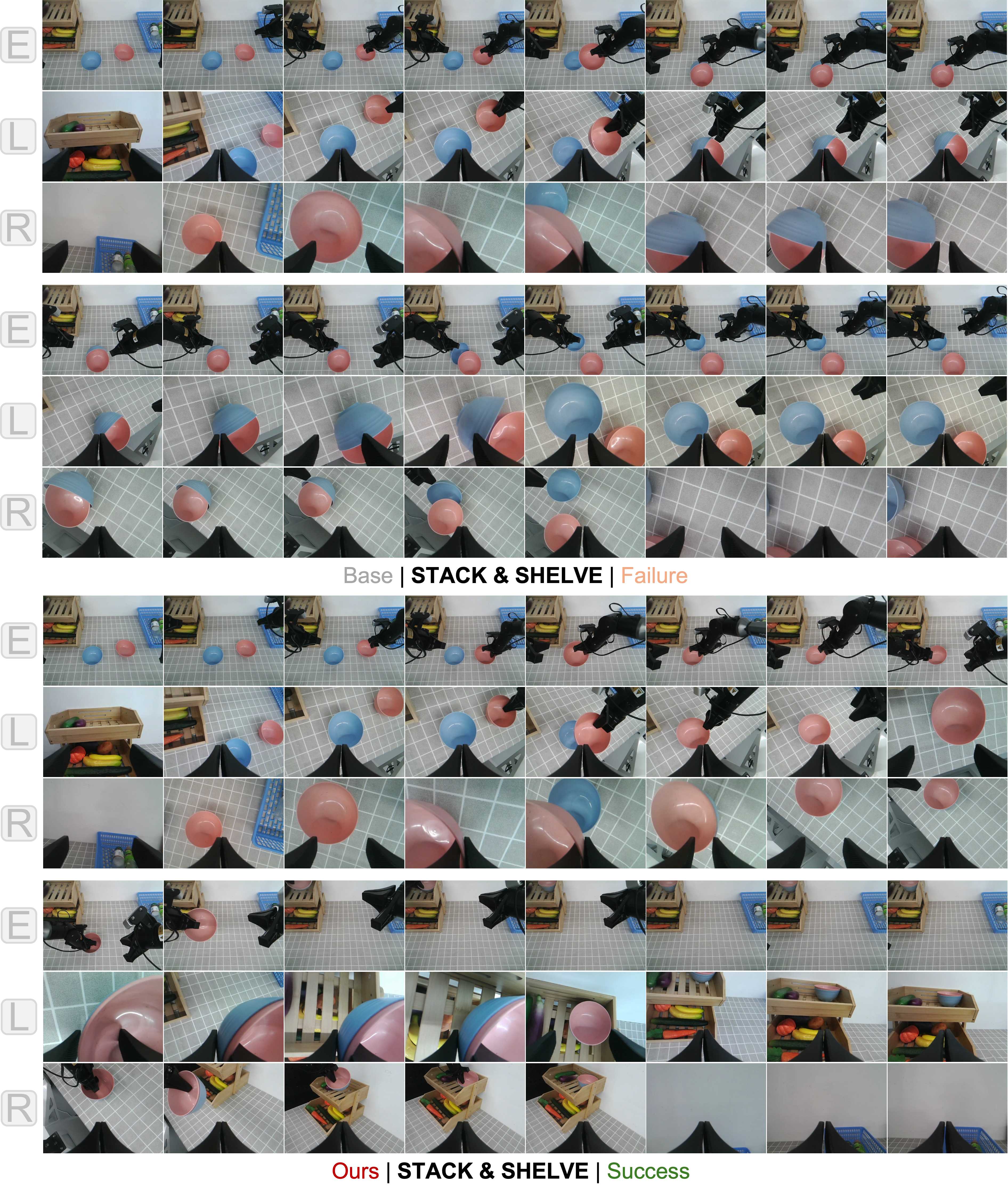}
  \caption{Stack \& Shelf: Case 2.}
  \Description{Real-world Stack and Shelf rollout snapshots for Case 2.}
  \label{fig:sup_realworld_stack_2}
\end{figure*}

\begin{figure*}[p]
  \centering
  \includegraphics[
    width=\textwidth,
    height=0.92\textheight,
    keepaspectratio
  ]{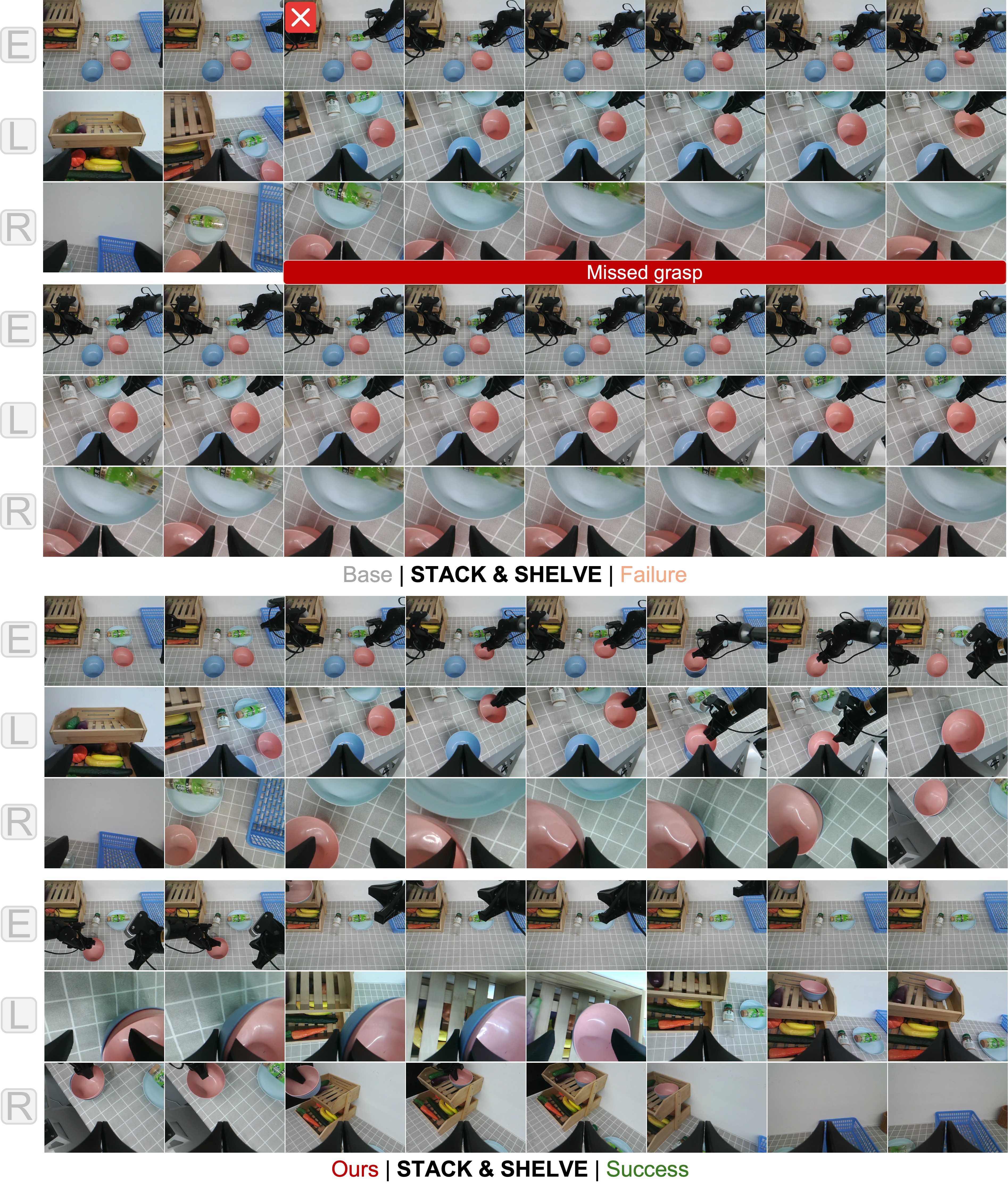}
  \caption{Stack \& Shelf: OOD Clutter.}
  \Description{Real-world Stack and Shelf rollout snapshots for the OOD clutter case.}
  \label{fig:sup_realworld_stack_ood}
\end{figure*}

\begin{figure*}[p]
  \centering
  \includegraphics[
    width=\textwidth,
    height=0.92\textheight,
    keepaspectratio
  ]{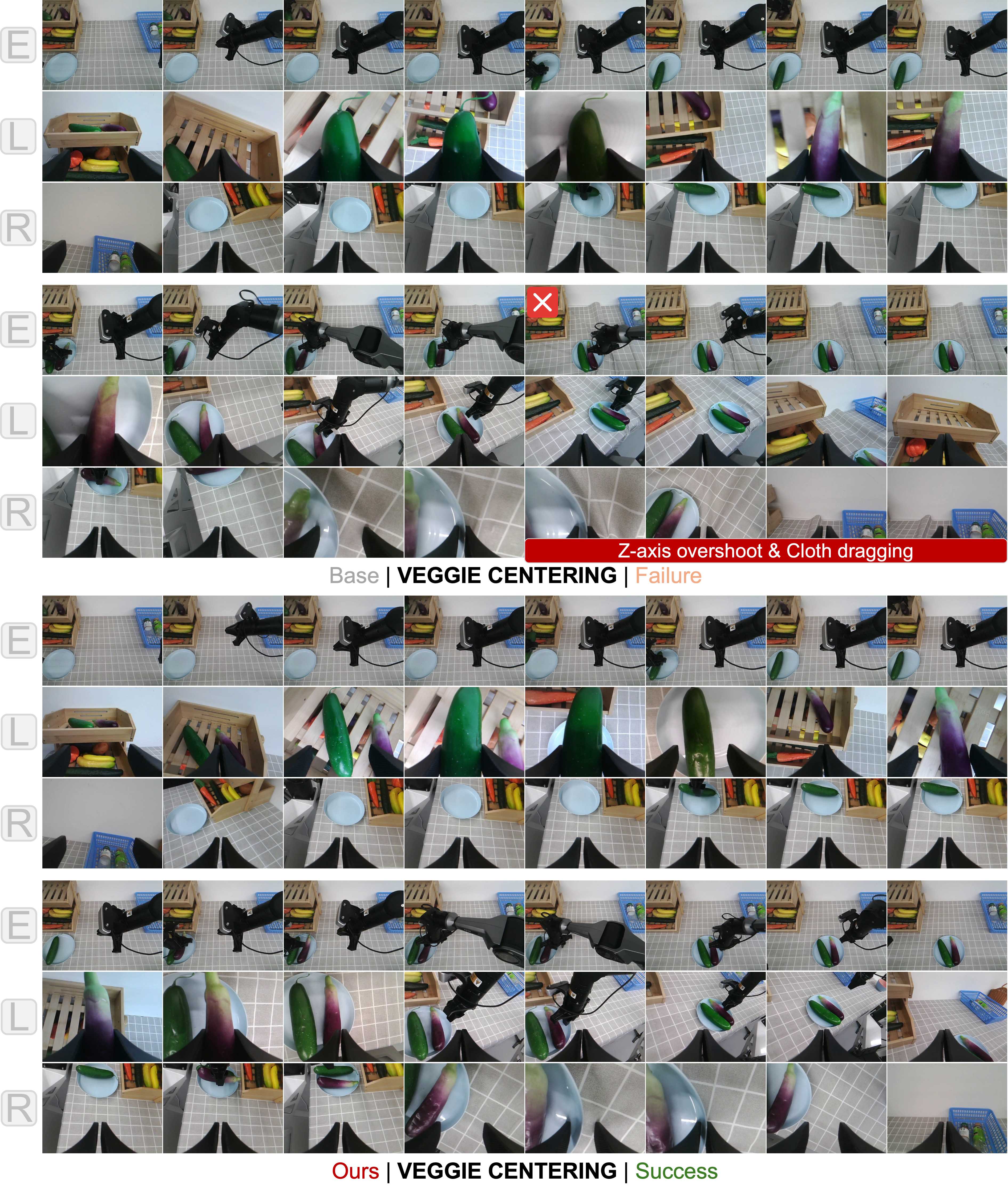}
  \caption{Veggie Centering: Case 1.}
  \Description{Real-world Veggie Centering rollout snapshots for Case 1.}
  \label{fig:sup_realworld_veggie_1}
\end{figure*}

\begin{figure*}[p]
  \centering
  \includegraphics[
    width=\textwidth,
    height=0.92\textheight,
    keepaspectratio
  ]{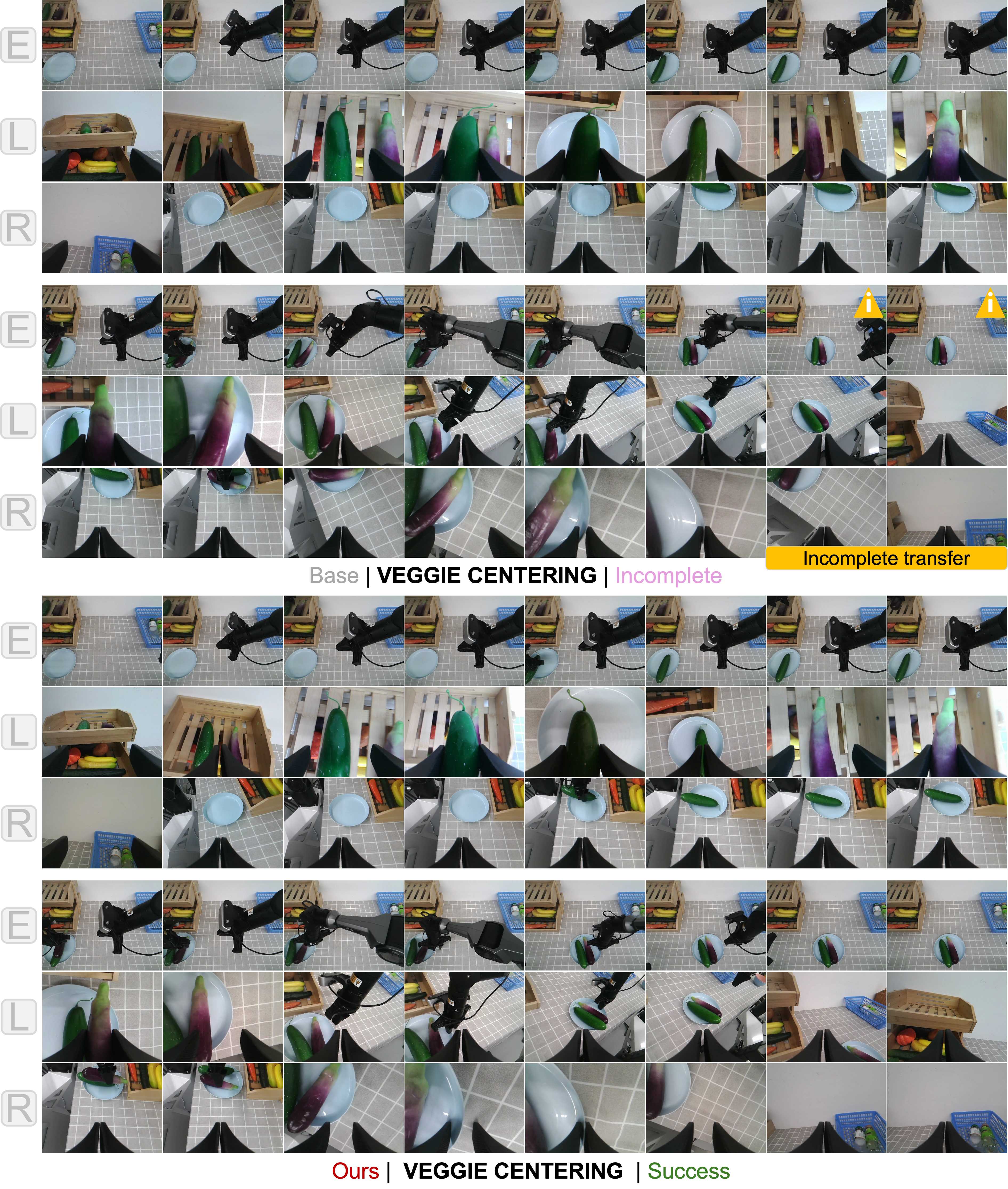}
  \caption{Veggie Centering: Case 2.}
  \Description{Real-world Veggie Centering rollout snapshots for Case 2.}
  \label{fig:sup_realworld_veggie_2}
\end{figure*}

\begin{figure*}[p]
  \centering
  \includegraphics[
    width=\textwidth,
    height=0.92\textheight,
    keepaspectratio
  ]{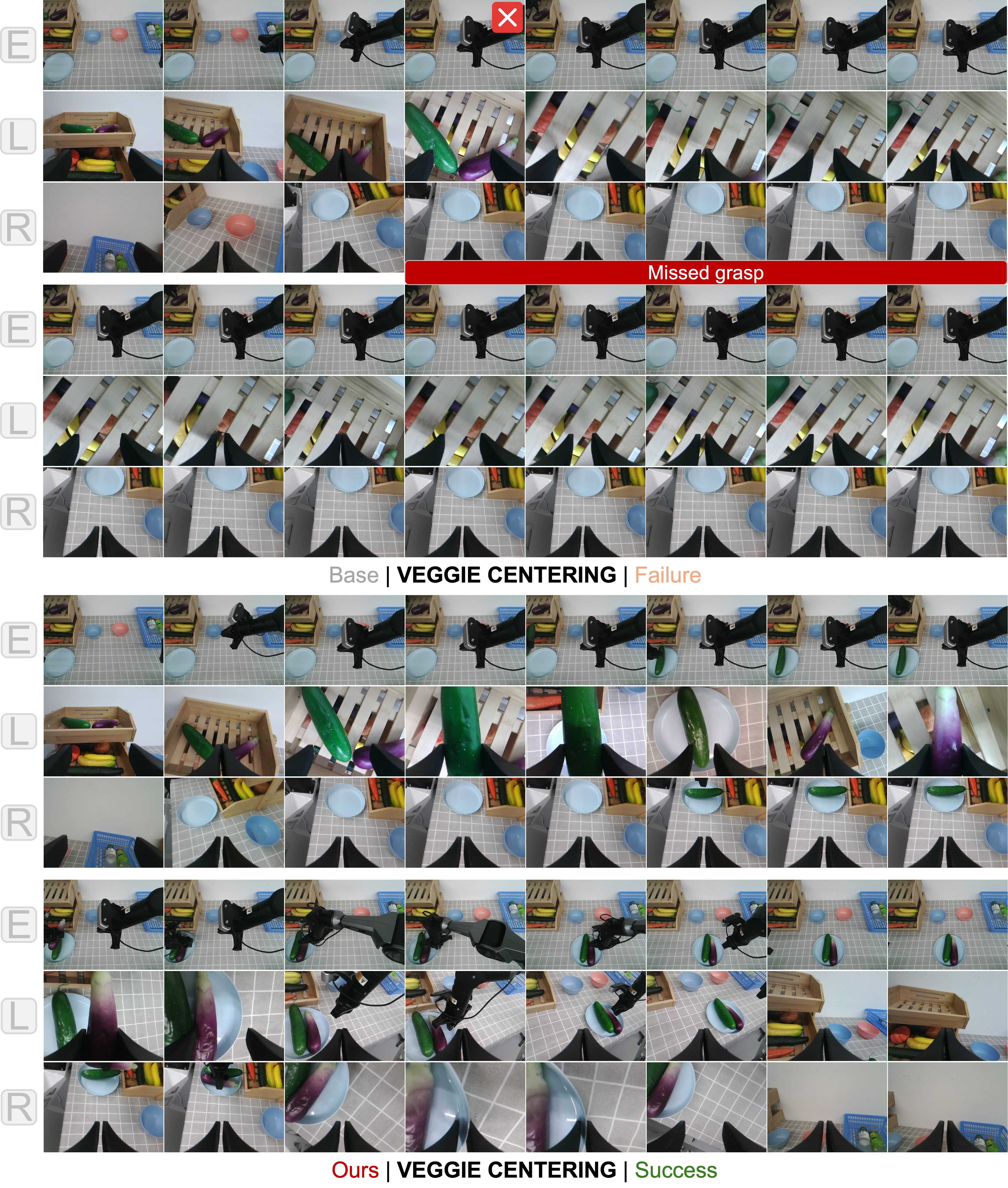}
  \caption{Veggie Centering: OOD Clutter.}
  \Description{Real-world Veggie Centering rollout snapshots for the OOD clutter case.}
  \label{fig:sup_realworld_veggie_ood}
\end{figure*}

Figs.~\ref{fig:sup_realworld_bottles_1}--\ref{fig:sup_realworld_veggie_ood} provide qualitative context for these aggregate rates. For \textbf{Bottle Cleanup}, Fig.~\ref{fig:sup_realworld_bottles_1} shows a clean case where the baseline reaches the bin with the brown-cap bottle but fails to release it, whereas M3 completes both disposal branches. Fig.~\ref{fig:sup_realworld_bottles_2} presents a second clean case in which both methods progress through the early pickups, yet only M3 enters the later transfer-and-disposal stage. Fig.~\ref{fig:sup_realworld_bottles_ood} depicts an OOD cluttered scenario where the baseline stalls at an early phase. For \textbf{Stack \& Shelf}, Fig.~\ref{fig:sup_realworld_stack_1} shows a clean case where the baseline narrowly completes the full task despite an imprecise bowl stack, whereas M3 maintains smoother execution across multiple stages. Fig.~\ref{fig:sup_realworld_stack_2} shows a harder clean case where the baseline fails to complete the later pickup-and-placement stages. Fig.~\ref{fig:sup_realworld_stack_ood} presents an OOD example in which the baseline misses the initial bowl grasp, whereas M3 completes the full stack-and-shelf sequence. For \textbf{Veggie Centering}, Fig.~\ref{fig:sup_realworld_veggie_1} shows a baseline failure at the final centering stage, with visible downward drift that perturbs the cloth. Fig.~\ref{fig:sup_realworld_veggie_2} presents a second clean case where the baseline reaches the transport stage but leaves the transfer incomplete. Fig.~\ref{fig:sup_realworld_veggie_ood} depicts an OOD example in which the baseline fails at an early grasp while M3 proceeds to completion. Across these cases, the qualitative differences align with the phase-level trends in Table~\ref{tab:realworld_phase_multiround}.

\begin{figure*}[t!]
  \centering
  \includegraphics[
    width=\textwidth,
    height=0.94\textheight,
    keepaspectratio        
  ]{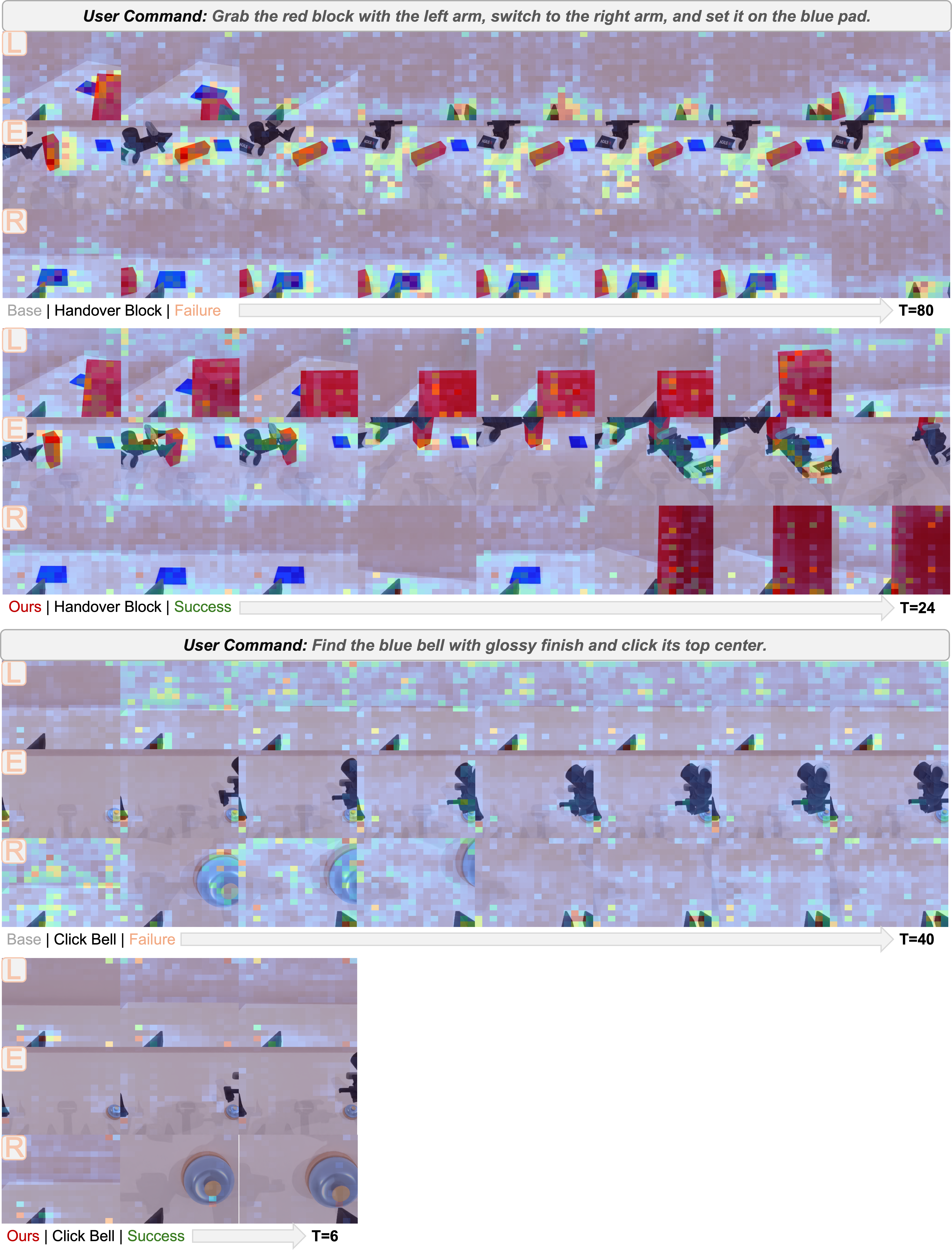}

\caption{Qualitative attention maps and rollout timelines comparing the baseline with M3 across representative bimanual tasks.}
  \Description{A large qualitative figure compares baseline and M3 attention maps and rollout timelines across multiple tasks. In the shown examples, M3 exhibits more concentrated attention on task-relevant regions and smoother execution sequences than the baseline.}

  \label{fig:task_heatmaps}
\end{figure*}

\noindent\textbf{Qualitative Attention Patterns.}
A core motivation of M3 is to reduce overfitting to spurious visual correlations, such as background textures or the robot's own gripper. These distractions are prevalent when fusing high-dimensional multi-view inputs. As illustrated in Fig.~\ref{fig:task_heatmaps}, the baseline policy often exhibits attention misalignment. In the \textit{Handover Block} and \textit{Click Bell} tasks, the baseline's attention maps are scattered across the table surface and irrelevant regions. This lack of focus is often accompanied by failures to localize the target object or the correct transfer point. In contrast, the M3-trained policy shows more concentrated attention in the shown examples. The heatmaps suggest that the model attends more consistently to task-relevant semantic regions, including the object to be grasped, the receiving hand, contact points, or the final placement pad. These qualitative patterns are consistent with the hypothesis that stochastic masking of arm views and queries during training may encourage the model to rely less on transient, salient pixel features. We also note that M3 does not completely remove attention misalignment in Clean2Rand. In several failure cases, distractor backgrounds, lighting variation, or visually salient arm regions still attract attention away from the true contact area. We attribute part of this gap to training policies on clean demonstrations while Clean2Rand evaluates them on scenes with amplified nuisance factors.

\noindent\textbf{Qualitative Execution Stability.}
Temporal execution differences are qualitatively consistent with the attention patterns discussed above. Fig.~\ref{fig:task_heatmaps} provides a timeline comparison between the baseline and M3. In the \textit{Click Bell} task (Fig.~\ref{fig:task_heatmaps}, bottom), the baseline policy struggles to ground the bell instruction, resulting in a prolonged search phase ($T=40\text{s}$) that ultimately ends in failure. In the shown rollout, the M3 policy appears to locate the target earlier and executes the click action in $T=6\text{s}$. Similarly, in the \textit{Handover Block} task (Fig.~\ref{fig:task_heatmaps}, top), the baseline policy inadvertently knocks over the target object during the interaction. This collision drives the scene into an out-of-distribution state where the policy becomes disoriented and unable to recover, causing the execution to stall at $T=80\text{s}$. By contrast, the M3 policy completes the successful handover and placement in $24\text{s}$ in the shown example. Across these examples, more focused attention coincides with more stable execution.

\begin{figure*}[t!]
  \centering
  \includegraphics[
    width=\textwidth,
    height=0.96\textheight,
    keepaspectratio        
  ]{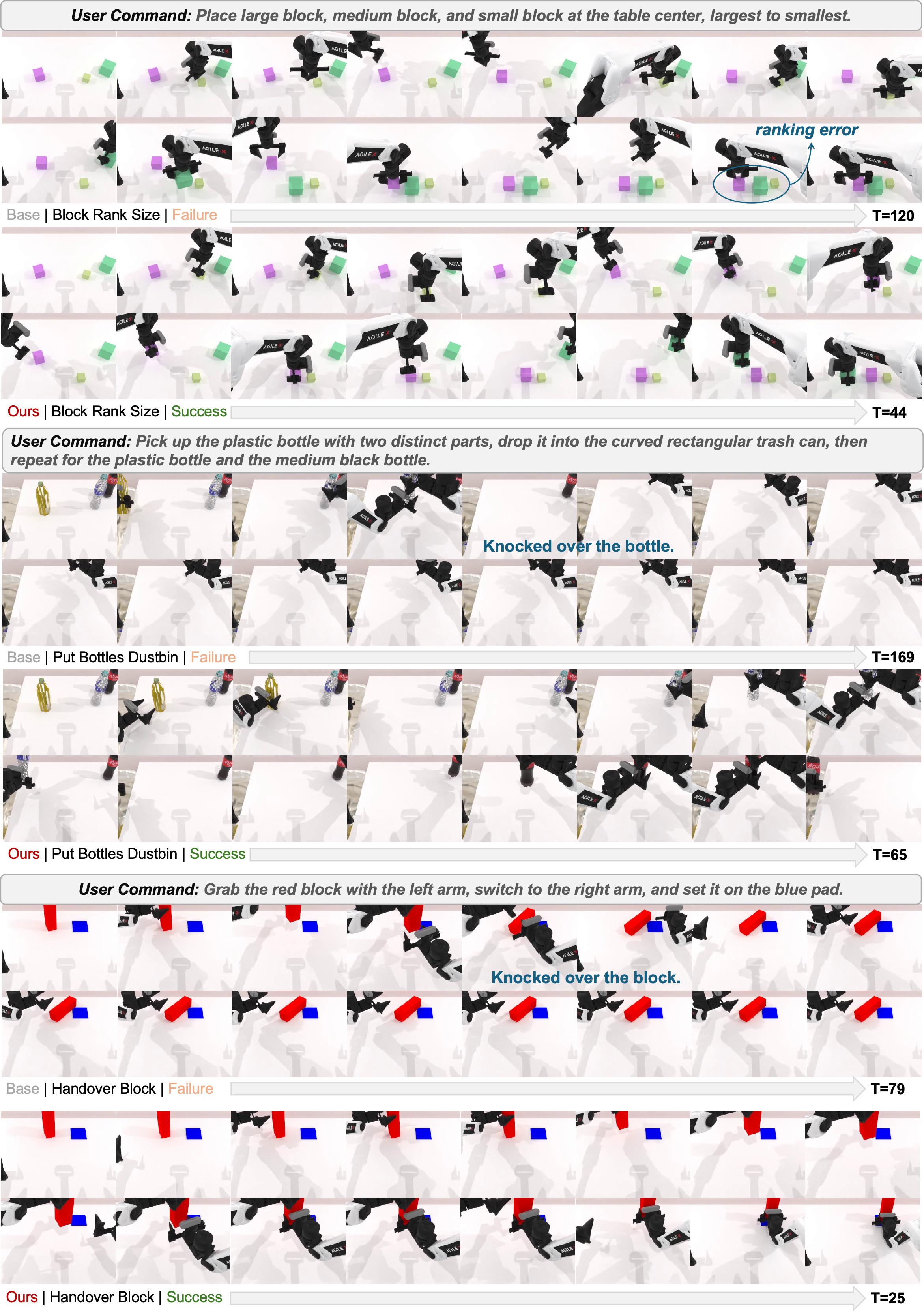}
\caption{Qualitative rollout comparisons on representative long-horizon bimanual manipulation tasks.}
  \Description{A rollout comparison across long-horizon bimanual tasks contrasts representative cases from the baseline and M3. The panels illustrate examples where M3 exhibits more stable multi-stage behavior and fewer accumulated errors relative to the baseline.}

  \label{fig:long_tasks}
\end{figure*}

\noindent\textbf{Qualitative Long-Horizon Rollout Comparisons.}
Bimanual manipulation often requires sequential reasoning, where early minor errors can compound into catastrophic failure. Fig.~\ref{fig:long_tasks} illustrates comparative rollouts for complex, long-horizon tasks. In the \textit{Block Rank Size} task (Fig.~\ref{fig:long_tasks}, top), which requires sorting objects by size, the baseline correctly grasps the first block but fails to maintain the logical ordering during placement, leading to a ranking error at $T=120\text{s}$. In the shown example, M3 successfully arranges the blocks from largest to smallest by $T=44\text{s}$. Furthermore, in the \textit{Put Bottles Dustbin} and \textit{Handover Block} tasks (Fig.~\ref{fig:long_tasks}, middle and bottom), the baseline often fails during contact-rich phases. In the \textit{Put Bottles Dustbin} task, the policy makes unstable contact during the manipulation process. These errors accumulate over time and result in the target bottle being knocked over, as observed in the final failure state at $T=169\text{s}$. In contrast, M3 completes these intricate sequences without disturbing the environment state in the shown rollouts. Overall, these examples are qualitatively consistent with improved long-horizon stability under M3.

\noindent\textbf{Discussion.} Collectively, these visualizations are consistent with the possibility that the dynamic visibility constraints introduced by M3 \textit{may function as a regularization mechanism}. By encouraging the model to extract complementary evidence from vision and language under partial observability, the strategy may reduce reliance on spurious correlations. We hypothesize that such stochastic occlusion may penalize reliance on salient but task-irrelevant features and encourage the model to anchor its decision-making on more durable semantic and geometric cues. These qualitative patterns may help explain the more focused attention maps, lower execution instability, and improved long-horizon performance observed in our experiments.

{\color{orange}\textbf{NOTE:}} \textbf{See Fig.~\ref{fig:task_heatmaps} for attention maps, Fig.~\ref{fig:long_tasks} for long-horizon rollouts, and Figs.~\ref{fig:sup_realworld_bottles_1}--\ref{fig:sup_realworld_veggie_ood} for real-world cases on the following pages.}

\clearpage
\clearpage
\clearpage
\clearpage
\clearpage
\clearpage
\clearpage
\clearpage
\clearpage
\clearpage

\end{document}